\documentclass[10pt,twocolumn,letterpaper]{article}

\usepackage[pagenumbers]{cvpr} 

\definecolor{cvprblue}{rgb}{0.21,0.49,0.74}
\usepackage[pagebackref,breaklinks,colorlinks,allcolors=cvprblue]{hyperref}

\def\confName{CVPR}
\def\confYear{2027}

\definecolor{darkyellow}{RGB}{180,140,0}
\definecolor{darkgreen}{RGB}{0,100,0}
\usepackage{multirow}
\newcommand{\namedataset}{CMU-Drive}
\newcommand{\namemethod}{V2V-VLA}

\title{\namedataset: Cooperative Multi-agent Unified Driving with Reasoning}

\author{Hsu-kuang Chiu$^1$\quad Chien-Yi Wang$^2$\quad Min-Hung Chen$^3$ \quad Deva Ramanan $^1$\quad John M. Dolan$^1$ \\
 Stephen F. Smith$^1$ \\ 
$\textsuperscript{\rm 1}$Carnegie Mellon University,\quad$\textsuperscript{\rm 2}$Amazon,\quad$\textsuperscript{\rm 3}$NVIDIA
}

\begin{document}


\twocolumn[{%
    \renewcommand\twocolumn[1][]{#1}%
    \maketitle

    \centering
    \vspace{-30pt}
    \includegraphics[width=0.95\textwidth]{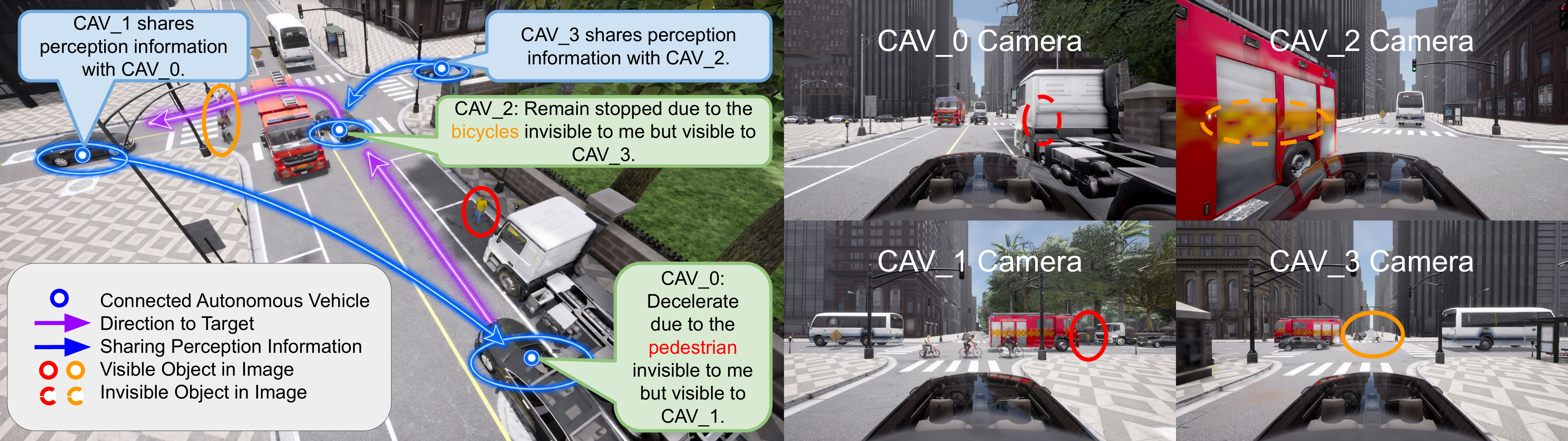}
    \vspace{-5pt}
    \captionsetup{type=figure,hypcap=false}
    \caption{
        Illustration of \namedataset~benchmark and \namemethod~model. Each connected autonomous vehicle (CAV) faces safety-critical driving scenarios. Each CAV is equipped with its own \namemethod~model, obtaining useful perception information from a selected communicating vehicle and generating driving actions, reasoning, communication policies, and future waypoints.
    }
    \label{fig:teaser}
    \vspace{0.5em}
}]

\begin{abstract}
Vision-Language-Action (VLA) models have recently achieved impressive performance for end-to-end autonomous driving, yet existing approaches are primarily designed for an individual single autonomous driving agent with limited support for cooperative perception, reasoning, and planning. We present \textbf{Cooperative Multi-agent Unified Driving with Reasoning (\namedataset)}, a closed-loop end-to-end benchmark for evaluating cooperative autonomous driving with multiple connected autonomous vehicles (CAVs) operating in safety-critical driving scenarios with background traffic participants. We further propose \textbf{Vehicle-to-Vehicle Vision-Language-Action (\namemethod)}, a cooperative VLA model that generates driving actions, language reasoning, communication policies, and future waypoints.
Experimental results demonstrate that our proposed model outperforms single-agent and multi-agent baselines, achieving relative improvements of $17.22\%$ in overall driving score and $27.28\%$ in overall success rate. Notably, in emergency braking scenarios, our model achieves relative improvements of $26.44\%$ in driving score and $57.13\%$ in success rate.
Our work provides a foundation for future research on multi-agent, closed-loop, end-to-end cooperative autonomous driving. 
Our code, benchmark, and model checkpoint will be publicly released to facilitate open-source research.
\end{abstract}    
\begin{table*}[tb]
  \caption{Benchmark attributes of \namedataset, in comparison to other driving benchmarks. \# Routes: the number of evaluation routes. \# CAVs: the number of connected autonomous vehicles per route. \# GPUs: the number of GPUs required to run the evaluation per route.
  }
  \vspace{-10pt}
  \label{tab:benchmark_comparison}
  \setlength{\tabcolsep}{1pt}
  \centering
  \small
  \begin{tabular}{l | cccc | cccc}
    \toprule
    Benchmark & Multi-Agent & End-to-End & Closed-Loop & Background Traffic & \# Routes & Max. \# CAVs & Avg. \# CAVs & \# GPUs \\
    \midrule
    Bench2Drive \cite{jia2024bench} & & \checkmark & \checkmark & \checkmark & 220 & 1 & 1 & \textbf{1} \\
    V2V-QA \cite{chiu2026v2vllm} & \checkmark & \checkmark & & \checkmark & 18 & 2 & 2 & \textbf{1} \\
    InterDrive~\cite{liu2025colmdriver} & \checkmark & \checkmark & \checkmark & & 46 & 8 & 3.80 & 3 \\
    MDrive~\cite{coscoy2026mdrive} & \checkmark & \checkmark & \checkmark & \checkmark & 225 & 8 & 3.36 & 3 \\ 
    \namedataset~(ours) & \checkmark & \checkmark & \checkmark & \checkmark & \textbf{250} & \textbf{16} & \textbf{6.63} & \textbf{1} \\
  \bottomrule
  \end{tabular}
  \vspace{-15pt}
\end{table*}

\section{Introduction}
\label{sec:introduction}

Vision-Language-Action (VLA) models~\cite{kim24openvla,rt12022rt1,rt22023rt2} have recently demonstrated strong capabilities for end-to-end autonomous driving by jointly performing visual perception, language reasoning, and vehicle control~\cite{renz2025simlingo,jia2024bench,nvidia2025alpamayo,li2025spacedrive,zhou2025autovla,hwang2024emma,tian2024DriveVLM,hu2025vla4ad,xing2024openemma}. However, existing VLA systems are primarily developed and evaluated in single-agent settings, where each autonomous vehicle makes driving decisions based on its own observations. In contrast, cooperative autonomous driving enables connected autonomous vehicles (CAVs) to share complementary perception information and collaboratively address safety-critical situations under partial observability. Extending VLA models to cooperative driving introduces new challenges in integrating observations from neighboring vehicles, reasoning about their relevance, and generating safe driving decisions.

Despite recent progress in cooperative autonomous driving, existing closed-loop end-to-end driving benchmarks~\cite{liu2025colmdriver,coscoy2026mdrive} typically involve relatively few cooperative vehicles and provide limited coverage of complex safety-critical scenarios involving diverse background traffic participants. Furthermore, existing cooperative driving methods focus mainly on resolving conflicting driving intentions through language-based negotiation~\cite{liu2025colmdriver}, rather than actively selecting complementary perception information from neighboring vehicles to improve driving decisions. These limitations motivate the development of a scalable cooperative driving benchmark and a VLA model that jointly performs cooperative perception, reasoning, communication, and planning.

To address these challenges, we introduce \textbf{Cooperative Multi-agent Unified Driving with Reasoning (\namedataset)}, a closed-loop, end-to-end cooperative driving benchmark, as illustrated in \cref{fig:teaser}. Built upon the single-agent Bench2Drive~\cite{jia2024bench}, \namedataset~preserves rich interactions with background traffic participants while extending each scenario to include 2 to 16 CAVs. \namedataset~also includes $30$ more evaluation routes with large vehicles, such as busses and trucks. This enables simultaneous interactions among cooperative vehicles, non-cooperative vehicles, pedestrians, and cyclists in diverse safety-critical driving scenarios. Moreover, through an efficient evaluation architecture, \namedataset~supports closed-loop evaluation of up to 16 cooperative driving agents using only a single GPU. 
\cref{tab:benchmark_comparison} summarizes the key attributes of \namedataset.

We also propose \textbf{Vehicle-to-Vehicle Vision-Language-Action (\namemethod)}, a cooperative VLA model that integrates cooperative perception, language reasoning, communication policy learning, and driving action generation. Unlike negotiation-based approaches~\cite{liu2025colmdriver} that require multiple rounds of language interaction and LLM inference, \namemethod~jointly generates driving actions, language reasoning, communication policies, and future waypoints. This enables each CAV to determine when communication is beneficial and which neighboring vehicle can provide complementary observations for safer driving decisions. Together, \namedataset~and \namemethod~establish a unified benchmark and baseline for studying reasoning-guided cooperative multi-agent, closed-loop, end-to-end autonomous driving.
\section{Related Work}
\label{sec:related_work}

\noindent\textbf{Autonomous Driving Benchmarks.}
Bench2Drive~\cite{jia2024bench} provides comprehensive closed-loop evaluation for single-agent end-to-end driving~\cite{hu2023uniad,jiang2023vad,chen2024vadv2,zhai2023admlp,wu2022trajectoryguided,jia2023thinktwice,jia2023driveadapter,shao2023lmdrive} across diverse safety-critical scenarios. Recent benchmarks extend evaluation to multi-agent settings. InterDrive~\cite{liu2025colmdriver} focuses on interactions among multiple autonomous vehicles without much consideration on background traffic or vulnerable road users, while MDrive~\cite{coscoy2026mdrive} further introduces agentically generated scenarios and incorporates V2X scenarios~\cite{zhou2024v2xpnp}. However, existing cooperative driving benchmarks involve relatively few autonomous agents and provide limited coverage of complex interactions with background traffic participants. In contrast, our proposed \namedataset~supports up to 16 CAVs interacting with various other traffic participants. And our evaluation can be conducted with only a single GPU, as compared in \cref{tab:benchmark_comparison}.

\noindent\textbf{Vision-Language-Action Models.}
Recent VLA models~\cite{renz2025simlingo,nvidia2025alpamayo,li2025spacedrive,zhou2025autovla,hwang2024emma,tian2024DriveVLM,hu2025vla4ad,xing2024openemma} integrate visual perception, language reasoning, and vehicle control for end-to-end autonomous driving. In particular, SimLingo~\cite{renz2025simlingo} unifies driving, vision-language understanding, and language-action alignment, achieving state-of-the-art performance in both Bench2Drive~\cite{jia2024bench} and CARLA~\cite{dosovitskiy2017carla} challenges. However, these methods primarily focus on single-agent driving without explicitly leveraging complementary observations from neighboring vehicles.

\noindent\textbf{Cooperative Autonomous Driving.}
Existing works investigate cooperative perception~\cite{chen2019fcooper,wang2020v2vnet,li2021disco,xu2022v2xvit,xu2022cobevt,xiang2023hmvit,li2022v2xsim,xu2022opencood,yu2022dair-v2x,xu2023v2v4real,xiang2024v2xreal,zimmer2024tumtraf,zhou2024v2xpnp,liu2020who2com,liu2020when2com,yu2023v2xseq,lu2023coalign,lu2024heal}, communication-efficient information sharing~\cite{hu2022where2comm,luo2025mixed,zhou2025turbotrain,zhao2025coopre,zhao2025quantv2x,cho2025cocoon}, and driving~\cite{liu2024codriving}. Recent multimodal LLM-based approaches~\cite{chiu2026v2vllm,chiu2026v2vgot,gao2025langcoop,you2024v2xvlm} further explore cooperative perception and language reasoning, while CoLMDriver~\cite{liu2025colmdriver} resolves conflicting driving intentions through multi-round language negotiation. Unlike these approaches, our proposed \namemethod~jointly learns driving actions, language reasoning, communication policies, and future waypoints to selectively acquire complementary perception information from neighboring CAVs.
\section{\namedataset: Cooperative Multi-agent Unified Driving with Reasoning Benchmark}
\label{sec:benchmark}

\begin{figure*}[!t]
        \centering
        \begin{subfigure}[t]{0.24\textwidth}
            \centering 
            \includegraphics[width=\textwidth]{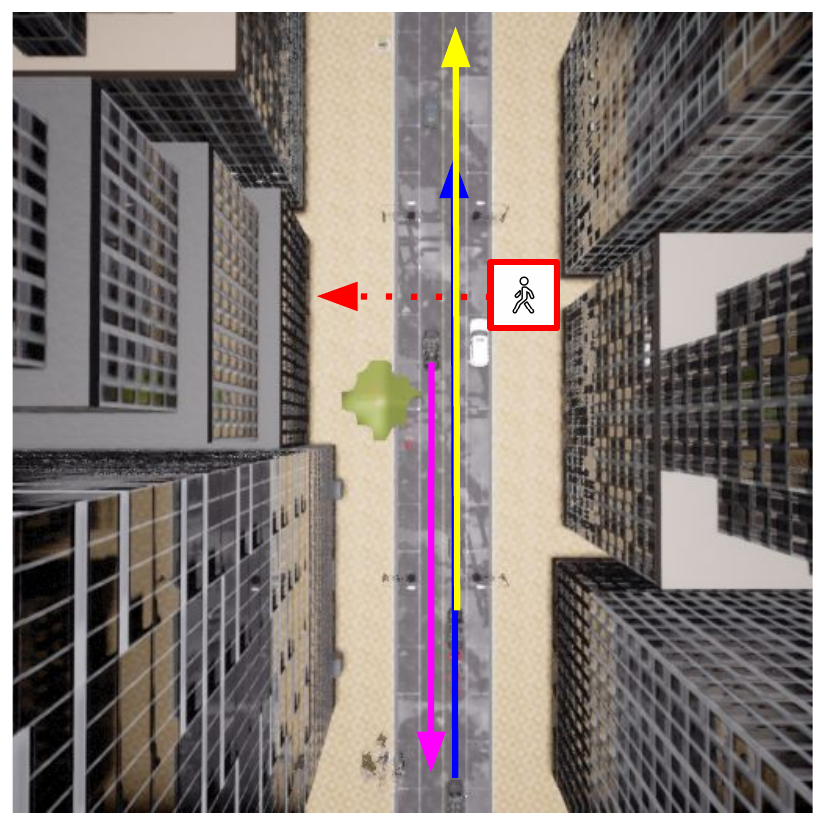}
            \caption[]%
            {{Pedestrian crossing from behind a parked vehicle. Total CAVs: 3.}}    
            \label{fig:route_extension_64}
        \end{subfigure}
        \hfill
        \begin{subfigure}[t]{0.24\textwidth}  
            \centering 
            \includegraphics[width=\textwidth]{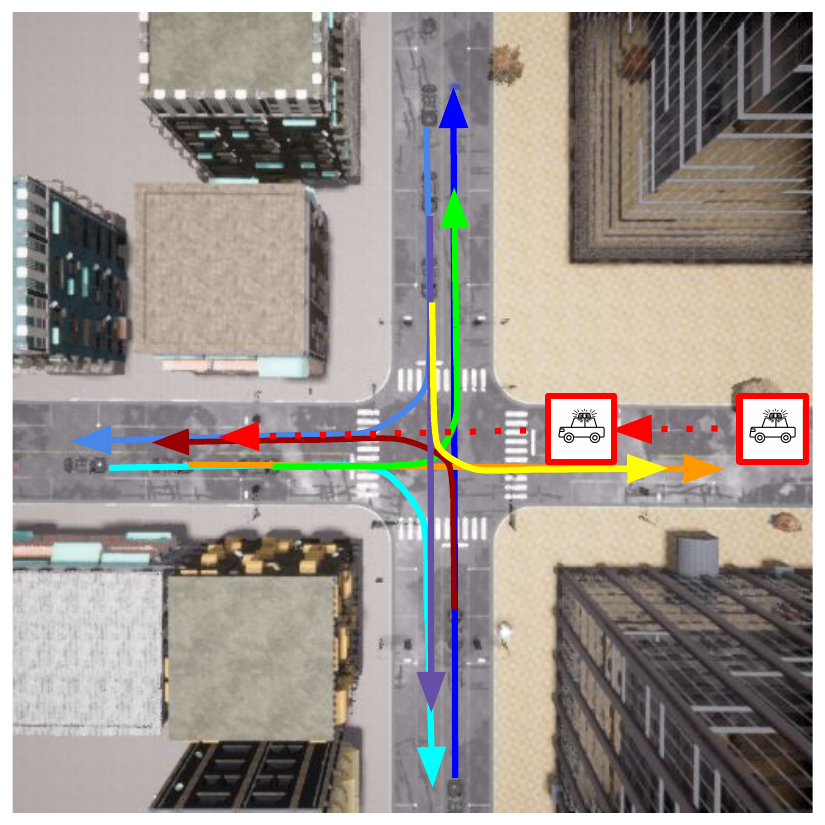}
            \caption[]%
            {{Emergency vehicles running a red light. Total CAVs: 8.}}    
            \label{fig:route_extension_93}
        \end{subfigure}
        \hfill
        \begin{subfigure}[t]{0.24\textwidth}
            \centering 
            \includegraphics[width=\textwidth]{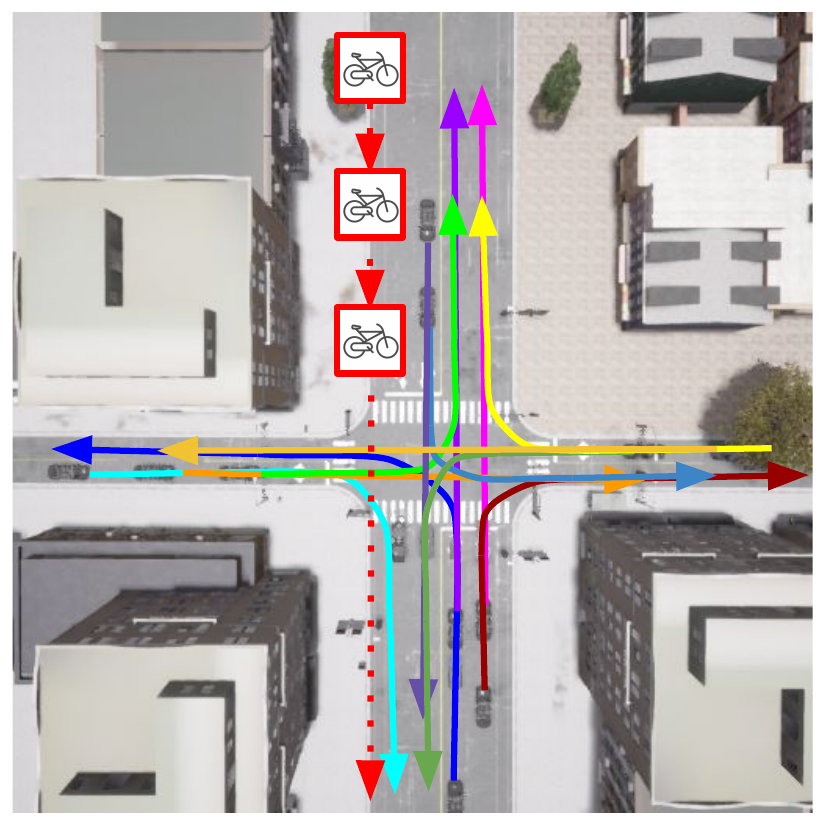}
            \caption[]%
            {{Bicycles crossing the road. Total CAVs: 12.}}    
            \label{fig:route_extension_58}
        \end{subfigure}
        \hfill
        \begin{subfigure}[t]{0.24\textwidth}
            \centering 
            \includegraphics[width=\textwidth]{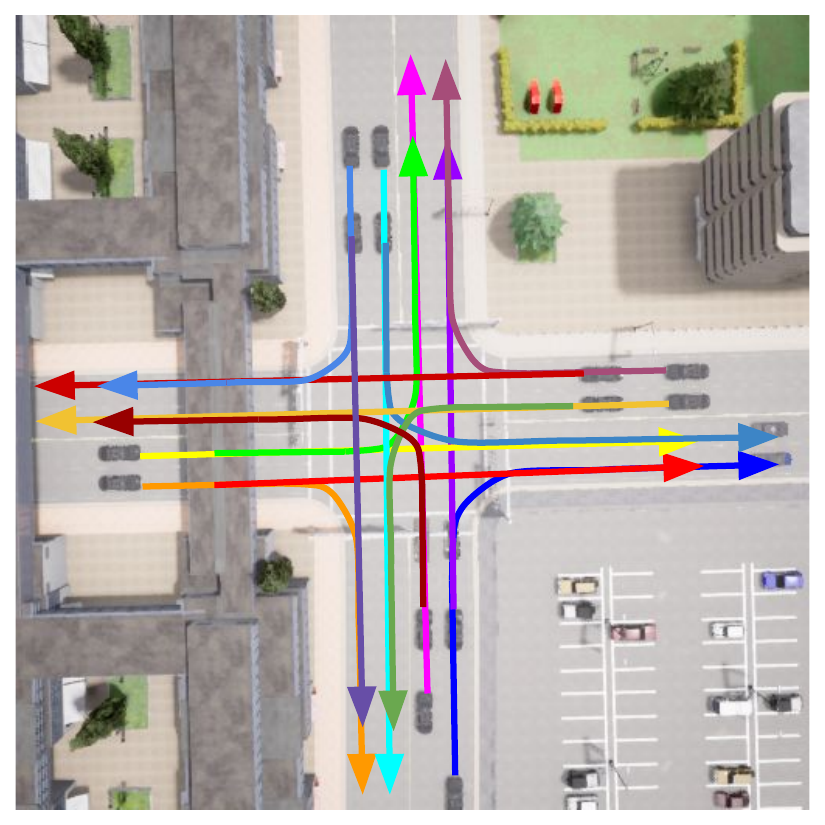}
            \caption[]%
            {{Vehicles in a busy intersection. Total CAVs: 16.}}    
            \label{fig:route_extension_196}
        \end{subfigure}

        \vspace{-10pt}
        \caption[]
        {
        Evaluation route samples in \namedataset~across different types of safety-critical scenarios and varying numbers of connected autonomous vehicles (CAVs). Each solid arrow represents a feasible path from the configured starting location to the destination of a CAV. Each dashed arrow represents the traffic flow of the configured background traffic participants within each safety-critical driving scenario. Each scenario also includes additional non-cooperative background vehicles driving nearby.
        } 
        \label{fig:route_extension}
        \vspace{-15pt}
\end{figure*}

\subsection{Benchmark Creation}

\namedataset~extends the widely adopted Bench2Drive~\cite{jia2024bench} benchmark by transforming its single-agent simulation and evaluation protocol into a cooperative multi-agent setting while preserving its diverse urban traffic scenarios. Specifically, \namedataset~extends the 44 types of scenarios from Bench2Drive~\cite{jia2024bench}, such as pedestrian crossing, emergency vehicles running a red light, and other safety-critical scenarios. Each scenario is instantiated under $5$ different weather and lighting conditions. In addition, ~\namedataset~further includes $6$ more scenario types and $30$ more evaluation routes with large vehicles, such as busses and trucks, resulting in a total of $50$ scenario types and $250$ evaluation routes. Details can be seen in the supplementary material.

Unlike Bench2Drive~\cite{jia2024bench}, where only one autonomous vehicle is simulated and evaluated in each route, \namedataset~ simultaneously simulates and evaluates 2 to 16 cooperative autonomous vehicles within the same closed-loop CARLA~\cite{dosovitskiy2017carla} environment, as shown in \cref{fig:route_extension}. Each cooperative vehicle is assigned an individual starting location and destination while sharing the same dynamic traffic environment.
The first cooperative vehicle follows the original Bench2Drive~\cite{jia2024bench} route configuration to preserve benchmark consistency and triggers the actions of the safety-critical background traffic participants following the same conditions configured in Bench2Drive~\cite{jia2024bench}. The starting locations and destinations of the additional cooperative vehicles are automatically generated according to the topology of the road of each scenario. For example, in junction scenarios, cooperative vehicles are initialized from different entry lanes with corresponding exit destinations, enabling diverse interactions among multiple autonomous agents and background traffic. 
To preserve the intended interactive traffic behaviors and multi-ability benchmarking from~\cite{jia2024bench},  
we avoid setting the starting location for the additional cooperative driving agents in the lanes that may interfere with the safety-critical scenario triggers defined in  Bench2Drive~\cite{jia2024bench}.
For example, in~\cref{fig:route_extension_93} where the scenario triggers emergency vehicles running a red light, we avoid introducing connected autonomous vehicles starting in the same lane as those emergency vehicles.
~\cref{fig:route_extension} illustrates more examples of \namedataset~evaluation routes.

\subsection{Multi-agent Evaluation Metrics}
We extend the single-agent evaluation protocol adopted by CarlaLeaderboard2.0~\cite{dosovitskiy2017carla},
Bench2Drive~\cite{jia2024bench}, and SimLingo~\cite{renz2025simlingo} to support multi-agent cooperative driving evaluation.
Specifically, for each cooperative autonomous driving agent $i$ in an evaluation route $j$ of \namedataset, we compute the agent-level route completion score $RC_i^j$ and the infraction score $IS_i^j$. The route completion $RC_i^j$ is defined as the percentage of the assigned route completed by agent $i$ in route $j$. The infraction score $IS_i^j$ is initialized with a base score $1.0$ and is multiplicatively penalized by a factor in the range of $[0.5, 0.8]$ whenever the agent commits a driving infraction, including colliding with pedestrian, cyclist, vehicle, or static objects, and running red lights or stop signs.
Details can be seen in the supplementary material.

For each route $j$, the route-level route completion score $RC^j$ is defined as the average agent-level route completion scores of all autonomous driving agents in route $j$, and the route-level infraction score $IS^j$ is defined as the product of all agent-level infraction scores.
\vspace{-5pt}
\begin{align}
  RC^j &= \frac{1}{N^j} \sum_{i=1}^{i=N^j} RC_i^j; IS^j = \prod_{i=1}^{i=N^j} IS_i^j
  \label{eq:route_level_route_completion_infraction}
\end{align}
, where $N^j$ is the number of connected autonomous vehicles in route $j$.
Next, the route-level driving score $DS^j$ is defined as the product of route completion score $RC^j$ and infraction score $IS^j$. The route-level success score $SS^j$ is set to $1$ if $DS^j = 100$, and 0 otherwise.
\vspace{-5pt}
\begin{align} 
  DS^j &= RC^j \cdot IS^j \\ 
  SS^j &= 
  \begin{cases}
    1, \text{if } DS^j = 100 \\
    0, \text{otherwise}
  \end{cases}
  \label{eq:route_level_driving_score_success_rate}
\end{align}
Finally, the overall driving score $DS$ is the average route-level driving score $DS^j$. The overall success rate $SR$ is the average route-level success score $SS^j$.
\vspace{-5pt}
\begin{align}
  DS &= \frac{1}{R} \sum_{j=1}^{j=R} DS^j; SR = \frac{1}{R} \sum_{j=1}^{j=R} SS^j
  \label{eq:overall_driving_sore_success_rate}
\end{align}
, where $R = 250$ is the total number of the evaluation routes in the \namedataset~benchmark.
\begin{figure*}[tb]
  \centering
  \includegraphics[width=0.9\textwidth]{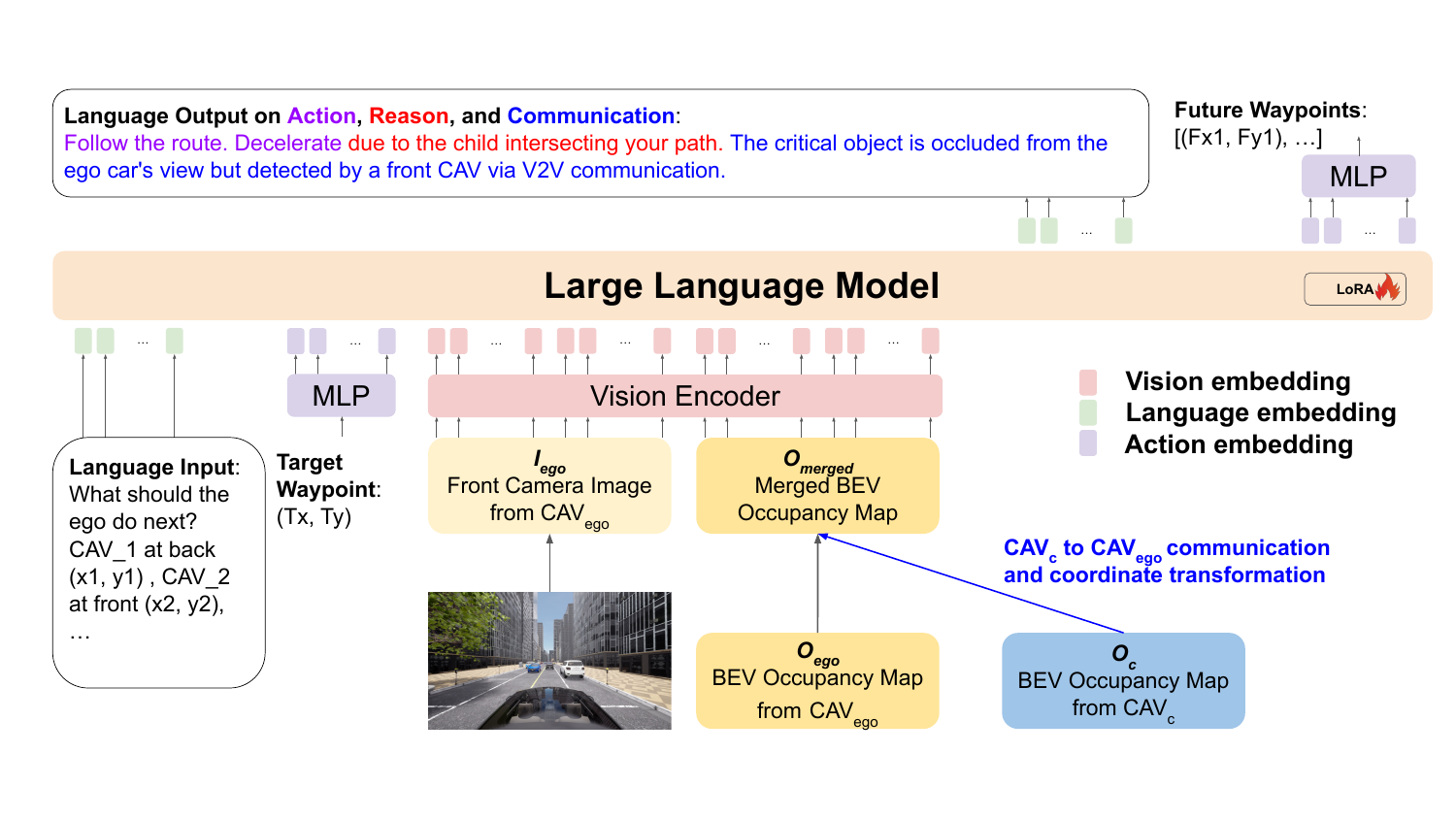}
  \vspace{-30pt}
  \caption{\namemethod~model architecture. The input consists of a language prompt, target waypoints, and vision tokens derived from the ego vehicle's front-camera image and a merged occupancy map from the ego and the selected communicating vehicle. The output comprises language outputs for the action, reasoning, and communication, alongside the future waypoints.
  }
  \label{fig:v2vvla_model}
  \vspace{-15pt}
\end{figure*}

\section{\namemethod: Vehicle-to-Vehicle Vision-Language-Action Model for Cooperative Autonomous Driving}
\label{sec:method}

We propose \namemethod, a new vehicle-to-vehicle vision-language-action model for cooperative autonomous driving, 
as shown in \cref{fig:v2vvla_model}. In each frame of the closed-loop simulation, each cooperative driving agent uses its own \namemethod~model to jointly generate driving actions, language reasoning, future waypoints, and an optional communication policy indicating whether another cooperative vehicle should be queried in the next time step.

\subsection{Vehicle-to-Vehicle Communication Algorithm} 

For an ego vehicle $CAV_{ego}$, its closest front cooperative autonomous vehicle is selected by default as its communicating vehicle $CAV_c$. However, we also train the \namemethod~model to generate the suggested communication target vehicle as part of the language output whenever there exists a good candidate. More specifically, if the \namemethod~model indicates that another cooperative vehicle $CAV_k$ is capable of detecting a critical object while the critical object is invisible to the ego vehicle $CAV_{ego}$, the ego vehicle $CAV_{ego}$ will use $CAV_k$ as the communicating vehicle $CAV_c$ in the next time step, instead of the default choice.

\subsection{Model Architecture}

\cref{fig:v2vvla_model} illustrates the overall architecture of \namemethod~model. Each cooperative autonomous driving agent is equipped with RGB cameras, an IMU, and GPS. At each time step, each cooperative autonomous driving agent feeds the language input, action input and vision input to its own \namemethod~model, and generates the language output on action, reasoning, communication, and future waypoints.

\textbf{Language input}: Language input $L_{input}$ encodes the prompt template that asks the model to describe the driving action that the ego vehicle $CAV_{ego}$ should take. The language input also includes the relative locations of other cooperative driving vehicles in the coordinate system of $CAV_{ego}$.

\textbf{Action input}: Action input $A_{input}$ encodes the target waypoint of $CAV_{ego}$ with a multi-layer-perceptron (MLP).

\textbf{Vision input}: Vision input $V_{input}$ includes two parts. The first part includes $CAV_{ego}$'s front RGB camera image $I_{ego}$. The second part is a merged camera-based bird-eye-view (BEV) occupancy map $O_{merged}$. Specifically, to create $O_{merged}$, both $CAV_{ego}$ and its selected communicating vehicle $CAV_c$ generate their individual 6-camera-based BEV occupancy map $O_{ego}$ and $O_c$. Subsequently, $O_c$ is warped to generate $O_{c -> ego}$ via coordinate transformation based on the ego vehicle pose $P_{ego}$ and the communicating vehicle pose $P_c$. Finally, $O_{merged}$ is generated by merging $O_{ego}$ and $O_{c -> ego}$ with the element-wise-or operator. $O_{merged}$ is then rendered as an RGB image, and is fed into a vision encoder, together with $I_{ego}$ to form the vision input $V_{input}$. 
\begin{align}
  O_{c -> ego} &= CoordinateTransform(O_c, P_{ego}, P_c) \\
  O_{merged} &= ElementwiseOr(O_{ego}, O_{c -> ego}) \\
  V_{input} &= VisionEncoder([I_{ego}, O_{merged}])
  \label{eq:vision_input}
\end{align} 
UniAD~\cite{hu2023uniad} is used to generate the 6-camera-based BEV occupancy maps, and InternViT~\cite{chen2024internvl} is used as the vision encoder.

\textbf{Vision-Language-Action model}: \namemethod~uses an LLM to take the language input, action input, and vision input and generate the language output $L_{output}$ and action output $A_{output}$. We use Qwen2~\cite{yang2024qwen2technicalreport} as the LLM.
\begin{align}
  L_{output}, A_{output} &= LLM(L_{input}, A_{input}, V_{input})
  \label{eq:v2vvla}
\end{align} 

\textbf{Language output}: Language output $L_{output}$ is generated auto-regressively, explaining the driving decision, and identifies safety-critical objects whenever necessary. \namemethod~also generates reasoning comments related to critical objects invisible to the ego vehicle and the suggested communication target vehicle $CAV_c$ whenever there exist.

\textbf{Action output}: We follow the same approach as in SimLingo~\cite{renz2025simlingo} to format the action output $A_{output}$. The action output embeddings are generated with trainable action-query embeddings and is conditioned on the model input and the language output. The action output embeddings are then used by a multi-layer-perceptron (MLP) to generate the future \textit{isochronous} and \textit{equidistant} waypoints. The waypoints are then used by PID-Controllers~\cite{emirler2014pid} to generate the control, including steer, throttle, and brake, for $CAV_{ego}$.

\subsection{Training}

\subsubsection{Training Data Collection}

First, we extend a subset of the single-agent training routes in SimLingo~\cite{renz2025simlingo} to build the multi-agent training routes for our \namedataset~ benchmark.
Then we collect the \namedataset~training data by using PDM-lite~\cite{Beibwenger2024PdmLite} as the expert driving model in each autonomous driving agent. PDM-lite~\cite{Beibwenger2024PdmLite} uses the ground-truth location and velocity of all objects in the CARLA simulation environment to make driving decisions.

At each time step, we save the future waypoints of the PDM-lite agent as the ground-truth annotation of the action output $A_{output}$. We use a rule-based approach, similar to DriveLM~\cite{sima2023drivelm} and SimLingo~\cite{renz2025simlingo}, to generate the suggested action and the reasoning parts of $L_{output}$.
Unlike previous works, we additionally annotate communication supervision. At each time step, we identify critical objects that influence the action of the expert driving agent. If such a critical object is invisible to the ego vehicle but observable by another cooperative vehicle, the annotation recommends that vehicle as the communication partner. 
Overall, we generate $182,328$ training samples for supervised fine-tuning.

\subsubsection{Training Details}
The training loss includes a smooth-L1 loss on the future waypoints in the action output $A_{output}$ and a cross-entropy loss on the language output $L_{output}$. We initialize the model weights from SimLingo~\cite{renz2025simlingo} and UniAD~\cite{hu2023uniad}. We fine-tune  the LLM with LoRA~\cite{hu2022lora} and freeze the UniAD~\cite{hu2023uniad} model. We also fine-tune all other trainable layers. We use 8 NVIDIA H100-80GB GPUs to train \namemethod~for $12$ epochs in 48 hours, with a batch size $8$. We follow SimLingo~\cite{renz2025simlingo} for setting all other hyperparameters.

\section{Experimental Results}
\label{sec:experiment}

\subsection{Quantitative Results}

\begin{table*}[tb]
  \caption{Quantitative experimental results in all evaluation routes of \namedataset~and the subset routes that evaluate driving model's skills of Merging, Overtaking, EmergencyBrake, GiveWay, and TrafficSign understanding. DS: Driving Score, SR: Success Rate.
  }
  \vspace{-10pt}
  \label{tab:quantitative_experimental_result}
  \small
  \setlength{\tabcolsep}{2pt}
  \centering
  \begin{tabular}{l | c | cc cc cc cc cc cc}
    \toprule
    \multirow{2}{*}{Method} & \multirow{2}{*}{Cooperation} & \multicolumn{2}{c}{All } & \multicolumn{2}{c}{Merging} & \multicolumn{2}{c}{Overtaking} & \multicolumn{2}{c}{EmergencyBrake} & \multicolumn{2}{c}{GiveWay} & \multicolumn{2}{c}{TrafficSign} \\

    & &
    \multicolumn{2}{c}{250 routes} & \multicolumn{2}{c}{85 routes} & \multicolumn{2}{c}{45 routes} & \multicolumn{2}{c}{85 routes} & \multicolumn{2}{c}{10 routes} & \multicolumn{2}{c}{105 routes} \\
    
    \cmidrule(lr){3-4} \cmidrule(lr){5-6} \cmidrule(lr){7-8} \cmidrule(lr){9-10} \cmidrule(lr){11-12} \cmidrule(lr){13-14}
    
    & &
    \multicolumn{1}{c}{DS $\uparrow$} & \multicolumn{1}{c}{SR (\%) $\uparrow$} & \multicolumn{1}{c}{DS $\uparrow$} & \multicolumn{1}{c}{SR (\%) $\uparrow$} & \multicolumn{1}{c}{DS $\uparrow$} & \multicolumn{1}{c}{SR (\%) $\uparrow$} & \multicolumn{1}{c}{DS $\uparrow$} & \multicolumn{1}{c}{SR (\%) $\uparrow$} & \multicolumn{1}{c}{DS $\uparrow$} & \multicolumn{1}{c}{SR (\%) $\uparrow$} & \multicolumn{1}{c}{DS $\uparrow$} & \multicolumn{1}{c}{SR (\%) $\uparrow$} \\
    
    \midrule
    SimLingo~\cite{renz2025simlingo} & $\times$ & 56.14
     & 30.80 & 45.80 & 20.00 & 68.75 & \textbf{44.44} & 55.94 & 32.94 & 77.60 & \textbf{40.00} & 42.38 & 19.05 \\ 
     
    V2V-LLM \cite{chiu2026v2vllm} & $\checkmark$ & 42.67 & 18.00 & 43.70 & 17.65 & 47.49 & 8.89 & 41.60 & 25.88 & 70.48 & \textbf{40.00} & 18.50 & 1.90 \\
    \hline
     
    \namemethod~(ours) & $\checkmark$ & \textbf{65.81} & \textbf{39.20} & \textbf{58.49} & \textbf{29.41} & \textbf{69.58} & 40.00 & \textbf{70.73} & \textbf{51.76} & \textbf{81.00} & \textbf{40.00} & \textbf{50.83} & \textbf{21.90}\\
    
  \bottomrule
  \end{tabular}
\end{table*}

\begin{table*}[tb]
  \caption{Ablation analysis in the sunny-day subset (50 routes) of the evaluation routes in \namedataset. DS: Driving Score, SR: Success Rate. CP: Communication Policy. \textcolor{red}{Red values}: relative percentage of performance changes in comparison to \namemethod.
  }
  \vspace{-10pt}
  \label{tab:ablation_study}
  \setlength{\tabcolsep}{5pt}
  \centering
  \begin{tabular}{l | l | c | c | c |cc }
    \toprule
    ID & Method & Cooperation & Reasoning & CP & DS $\uparrow$ & SR (\%) $\uparrow$ \\ 
    \midrule
    
    1 & V2V-VLA w/o cooperation & $\times$ & $\checkmark$ & $\times$ & 65.84 (\textcolor{red}{-5.54\%}) & 34.00 (\textcolor{red}{-15.00\%}) \\ 
    2 & V2V-VLA sharing image & camera image & $\checkmark$ & reasoning & 53.55 (\textcolor{red}{-23.17\%}) & 26.00 (\textcolor{red}{-35.00\%}) \\
    3 & V2V-VLA w/o reasoning & BEV occupancy map & $\times$ & random & 61.09 (\textcolor{red}{-12.36\%}) & 26.00 (\textcolor{red}{-35.00\%}) \\
    4 & V2V-VLA w/ random CP & BEV occupancy map & $\checkmark$ & random & 64.56 (\textcolor{red}{-7.39\%}) & 34.00 (\textcolor{red}{-15.00\%}) \\

    \hline 
    5 & \namemethod~(ours) & BEV occupancy map & $\checkmark$ & reasoning & \textbf{69.71} & \textbf{40.00} \\ 
  \bottomrule
  \end{tabular}
  \vspace{-10pt}
\end{table*}

\cref{tab:quantitative_experimental_result} summarizes the quantitative results across all $250$ evaluation routes in \namedataset, as well as subsets of routes designed to evaluate specific driving skills, including \textit{Merging}, \textit{Overtaking}, \textit{EmergencyBrake}, \textit{GiveWay}, and \textit{TrafficSign} understanding. Each evaluation route may assess one or more driving skills, as detailed in the supplementary material.
We compare \namemethod~against two baselines: single-agent SimLingo~\cite{renz2025simlingo} and multi-agent V2V-LLM~\cite{chiu2026v2vllm}.

As shown in \cref{tab:quantitative_experimental_result}, \namemethod~outperforms both baseline methods in driving score ($DS$) and success rate ($SR$) across \namedataset~ in most metrics. Specifically, compared with SimLingo~\cite{renz2025simlingo}, \namemethod~achieves relative improvements of 17.22\% in $DS$ and 27.28\% in $SR$, demonstrating the benefits of V2V cooperation. The skill-specific results further reveal that \namemethod~achieves its largest performance gains in the \textit{EmergencyBrake} category, with relative improvements of 26.44\% in $DS$ and 57.13\% in $SR$ over SimLingo~\cite{renz2025simlingo}. This finding highlights the particular importance of V2V cooperation in safety-critical driving scenarios. Scenarios in the \textit{EmergencyBrake} category typically involve situations in which vehicles, pedestrians, or cyclists initially occluded from the ego vehicle suddenly emerge from occluded regions and enter the ego vehicle's planned path, potentially creating imminent collision risks. By exchanging perception information with cooperative vehicles, \namemethod~can enable the ego vehicle to brake even when the critical objects are still invisible to the ego car's front camera image and avoid potential collisions. Across the other skill categories, \namemethod~also improves driving performance in most cases, demonstrating the broad benefits of cooperative perception across diverse scenarios.

In addition, \namemethod~significantly outperforms V2V-LLM~\cite{chiu2026v2vllm} across the evaluation routes in \namedataset. A key architectural difference between V2V-LLM and VLA-based driving models such as SimLingo~\cite{renz2025simlingo} and \namemethod~lies in how future waypoints are generated. Unlike these VLA-based methods, V2V-LLM does not employ dedicated action embeddings; instead, it directly uses the output language tokens to generate future waypoints. This design may limit the model's ability to generate smooth and physically plausible future trajectories, especially in scenarios require making turns or changing lanes.

\subsection{Ablation Analysis}
We conducted an ablation study on the sunny-day subset of \namedataset, which comprises 50 evaluation routes
in sunny weather. \cref{tab:ablation_study} summarizes the results, demonstrating the contributions of four key components of the \namemethod~architecture: vehicle-to-vehicle (V2V) cooperation (\ref{sec:ablation_cooperation}), BEV occupancy maps (\ref{sec:ablation_bev}), chain-of-thought (CoT) reasoning (\ref{sec:ablation_reasoning}), and the learning-based communication policy (\ref{sec:ablation_communication_policy}).

\subsubsection{V2V Cooperation}
\label{sec:ablation_cooperation}
We first investigated the impact of V2V cooperation by evaluating a variant of \namemethod~in which each agent relies solely on its own camera image and camera-based BEV occupancy map, without incorporating information from cooperative vehicles. As shown in row $1$ of \cref{tab:ablation_study}, the removal of the V2V cooperation reduces the driving score ($DS$) by 5.54\% and the success rate ($SR$) by 15.00\% in relative terms. These results demonstrate the importance of V2V cooperation in safety-critical driving scenarios.

\subsubsection{BEV Occupancy Map}
\label{sec:ablation_bev}
Next, we investigated the effectiveness of sharing BEV occupancy maps during V2V cooperation. To this end, we train a variant of \namemethod~that shares camera images from cooperative vehicles instead of their BEV occupancy maps. As shown in row $2$ of \cref{tab:ablation_study}, this variant exhibits substantial performance degradation, with relative decreases of 23.17\% in $DS$ and 35.00\% in $SR$.

We attribute this performance gap primarily to the varying relative poses between the ego vehicle's front-camera image and the camera image received from a cooperative vehicle. Because the relative pose between the two vehicles can change at every time step, the spatial relationship between these visual inputs is inherently dynamic, making it more challenging for the VLA model to learn effective spatial associations between the two visual inputs. In contrast, \namemethod~transforms the cooperative vehicle's BEV occupancy map into the ego vehicle's BEV coordinate system before merging the two maps. This alignment provides a consistent spatial reference for integrating the camera image and the BEV occupancy map information across time steps.

\subsubsection{Chain-of-Thought Reasoning}
\label{sec:ablation_reasoning}
We then evaluated the contribution of chain-of-thought (CoT) reasoning by training a variant that directly predicts future waypoints without first generating language outputs describing the driving action, reasoning process, and communication policy. As shown in row $3$ of \cref{tab:ablation_study}, eliminating this intermediate stage of language-generation reduces $DS$ by 12.36\% and $SR$ by 35.00\% in relative terms. These results suggest that the generation of intermediate language output contributes to improved driving performance. 

\begin{figure*}[!t]
        \centering
        \begin{subfigure}[t]{0.33\textwidth}
            \centering 
            \includegraphics[width=0.99\textwidth]{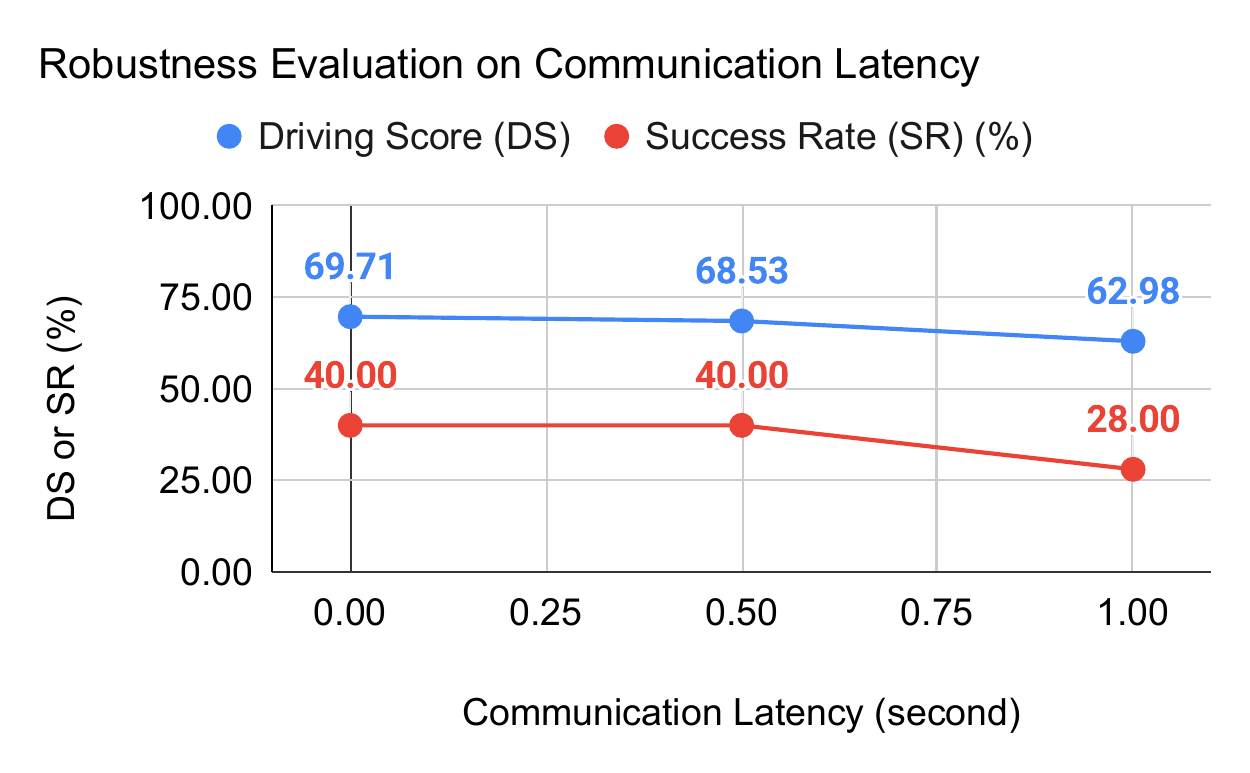}
            \vspace{-20pt}
            \caption{Robustness evaluation on latency.
            }
            \label{fig:latency}
        \end{subfigure}
        \hfill
        \begin{subfigure}[t]{0.33\textwidth}  
            \centering 
            \includegraphics[width=0.99\textwidth]{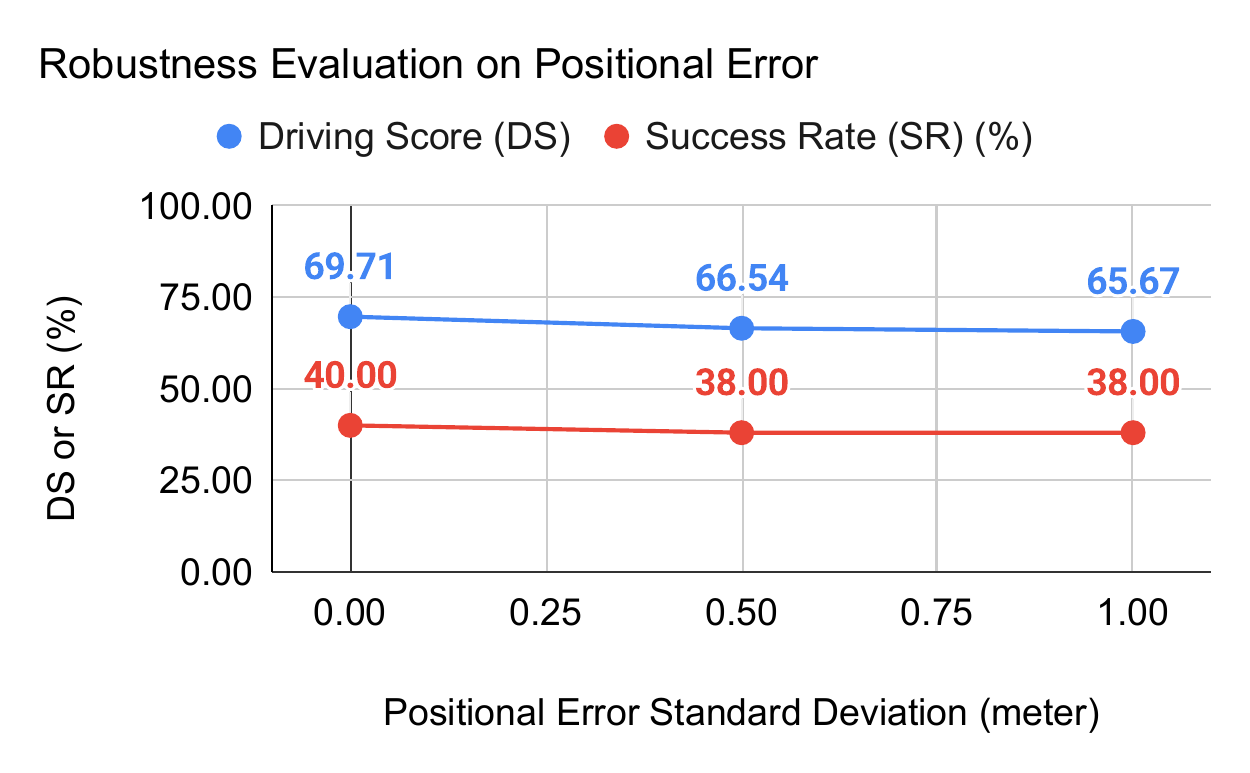}
            \vspace{-20pt}
            \caption{Robustness evaluation on positional error.
            }
            \label{fig:positional_error}
        \end{subfigure}
        \hfill
        \begin{subfigure}[t]{0.33\textwidth}
            \centering 
            \includegraphics[width=0.99\textwidth]{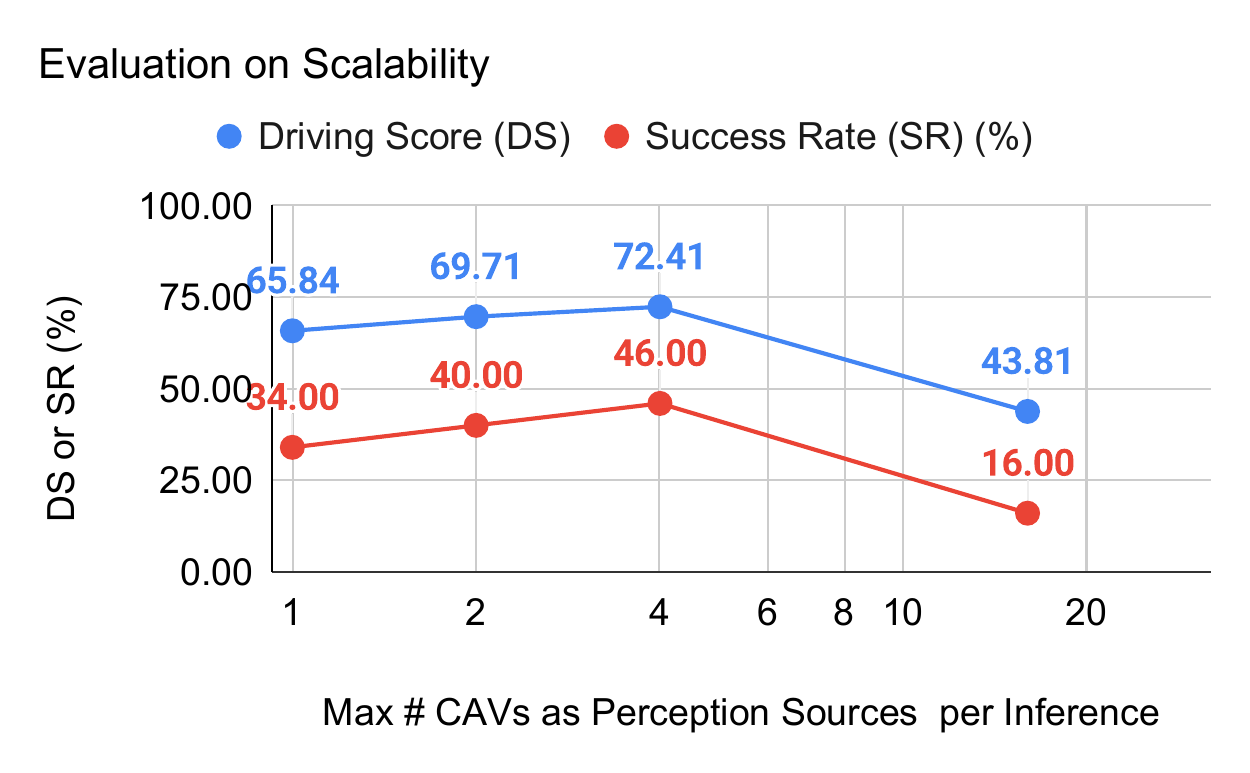}
            \vspace{-20pt}
            \caption{Evaluation on scalability.
            }
            \label{fig:scalability}
        \end{subfigure}
        
        \vspace{-10pt}
        \caption[]
        {
        Robustness evaluation on latency, positional error, and scalability 
        in the sunny-day subset (50 routes) of \namedataset.
        } 
        \label{fig:robustness_evaluation_scalability}
        \vspace{-15pt}
\end{figure*}

\subsubsection{Learning-based Communication Policy}
\label{sec:ablation_communication_policy}
Finally, we investigated the effectiveness of the learning-based communication policy. During inference, \namemethod~selects a cooperative vehicle based on the language output generated at the previous time step. For comparison, we evaluate a variant that randomly selects a cooperative vehicle at each time step. As shown in row $4$ of \cref{tab:ablation_study}, random selection reduces $DS$ by 7.39\% and $SR$ by 15.00\% in relative terms. These results indicate that the learning-based communication policy selects better vehicle pairings than random selection, allowing the ego vehicle to obtain more relevant complementary perception information and make better-informed driving decisions in safety-critical scenarios.

\subsection{Robustness to Communication Latency}
We evaluated the robustness of \namemethod~on communication latency on the same sunny-day subset of \namedataset. We used the BEV occupancy map of the selected cooperative vehicle captured $0.5$ or $1$ second ago and merged it with the occupancy map of the ego vehicle in the current time step. As shown in \cref{fig:latency}, driving performance gradually degrades as communication latency increases.


\subsection{Robustness to Positional Error}
We evaluated the robustness of \namemethod~on positional error on the  same sunny-day subset. We introduced Gaussian noise with standard deviations of $0.5$ or $1$ meters on the relative pose between the ego vehicle and its selected communicative vehicle. As shown in \cref{fig:positional_error},  
driving performance gradually degrades as the positional error increases.


\subsection{Scalability}
We evaluated the scalability of \namemethod~on the same sunny-day subset by developing alternative model configurations that aggregate BEV occupancy maps from multiple communicating CAVs into the second perception input channel. As shown in \cref{fig:scalability}, increasing the maximum number of CAVs as perception sources per inference from $2$ to $4$ further improves driving performance. However, incorporating perception information from all up to $16$ CAVs into each inference leads to significant performance degradation. This degradation may be attributed to the inclusion of less relevant or less accurate perception information from distant CAVs, which can introduce noise 
and negatively affect driving decisions. 


\subsection{Communication Cost Analysis}
In our experimental setting with \(N\ \leq 16\) connected autonomous vehicles (CAVs), denoted by \(\mathcal{C} = \{\mathrm{CAV}_i \mid i \in \{1,\ldots,N\}\}\), each ego vehicle \(\mathrm{CAV}_{\mathrm{ego}}\) receives, at each frame, a pose matrix \(P_i \in \mathbb{R}^{4 \times 4}\) from each of the other \(N-1\) CAVs and a binary BEV occupancy map \(O_c \in \mathbb{R}^{200 \times 200 \times 1}\) from a single selected communicating vehicle \(\mathrm{CAV}_c\). Assuming that each floating-point value is represented using 32 bits and each occupancy cell is encoded using 1 bit, 
the communication receiving cost of the pose matrices and the BEV occupancy map \(4 \times 4 \times 32 \times (N-1) + 200 \times 200 \times 1 \times 1 = 512(N-1) + 40{,}000 \) bits per frame.
Given the frame rate $F=20$ frames per second and the number of connected autonomous vehicles is bounded by \(N \leq 16\),
the total communication receiving cost is \(\left(512(N-1)+40{,}000\right)\times F = \left(512(16-1)+40{,}000\right)\times 20 = 953{,}600\) bits per second, or \(0.95\) Mb/s.



On the other hand, each CAV broadcasts its pose matrix to all other CAVs at each frame, while its BEV occupancy map is transmitted only when requested by at least one other CAV. Since existing V2V communication technologies, such as Dedicated Short-Range Communications (DSRC), C-V2X, and 5G NR-V2X~\cite{lee2013empirical,naik2019ieee,garcia2021tutorial}, support broadcast or groupcast communication, a CAV can transmit its information
once to serve multiple requesting CAVs, making the transmission cost independent of the number of receivers. 
Therefore, the maximum communication transmitting cost per CAV is \((4 \times 4 \times 32 + 200 \times 200 \times 1)\times 20 = 810{,}240\) bits per second, or \(0.81\) Mb/s. 

Furthermore, in the worst case where all \(N=16\) CAVs broadcast their pose matrices and BEV occupancy maps at every frame, the maximum aggregate transmitted communication payload rate is \((4 \times 4 \times 32 + 200 \times 200 \times 1)\times 20 \times 16 = 12{,}963{,}840\) bits per second, or approximately \(12.96\) Mb/s, which is below the maximum nominal physical-layer data rate of 27 Mb/s supported by DSRC~\cite{lee2013empirical,naik2019ieee,garcia2021tutorial}.


\begin{figure*}[!t]
        \centering
        \begin{subfigure}[t]{0.33\textwidth}
            \centering 
            \includegraphics[width=0.99\textwidth]{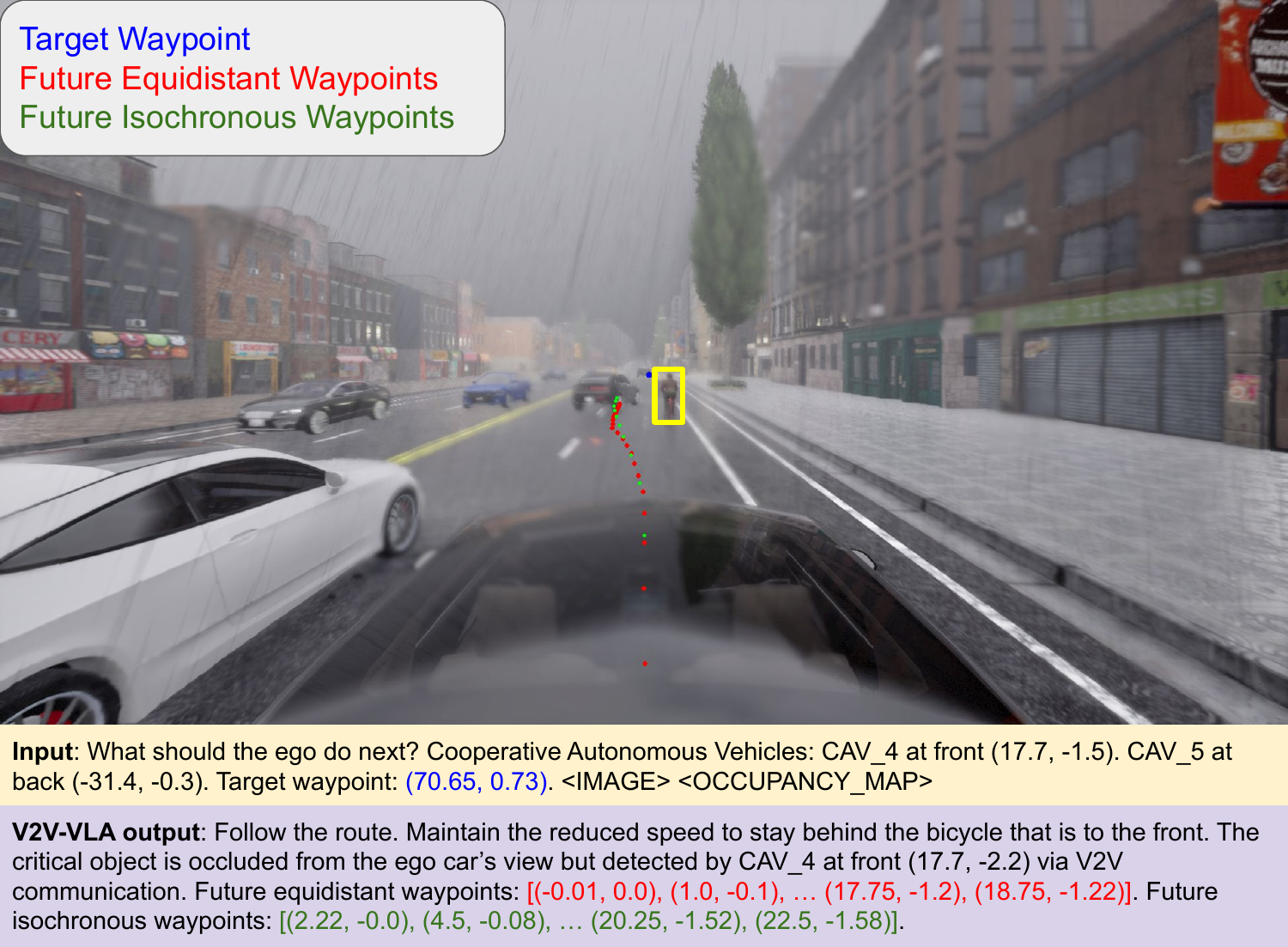}
            \caption{\namemethod~(ours)
            }
            \label{fig:qualitative_v2vvla_1}
        \end{subfigure}
        \hfill
        \begin{subfigure}[t]{0.33\textwidth}  
            \centering 
            \includegraphics[width=0.99\textwidth]{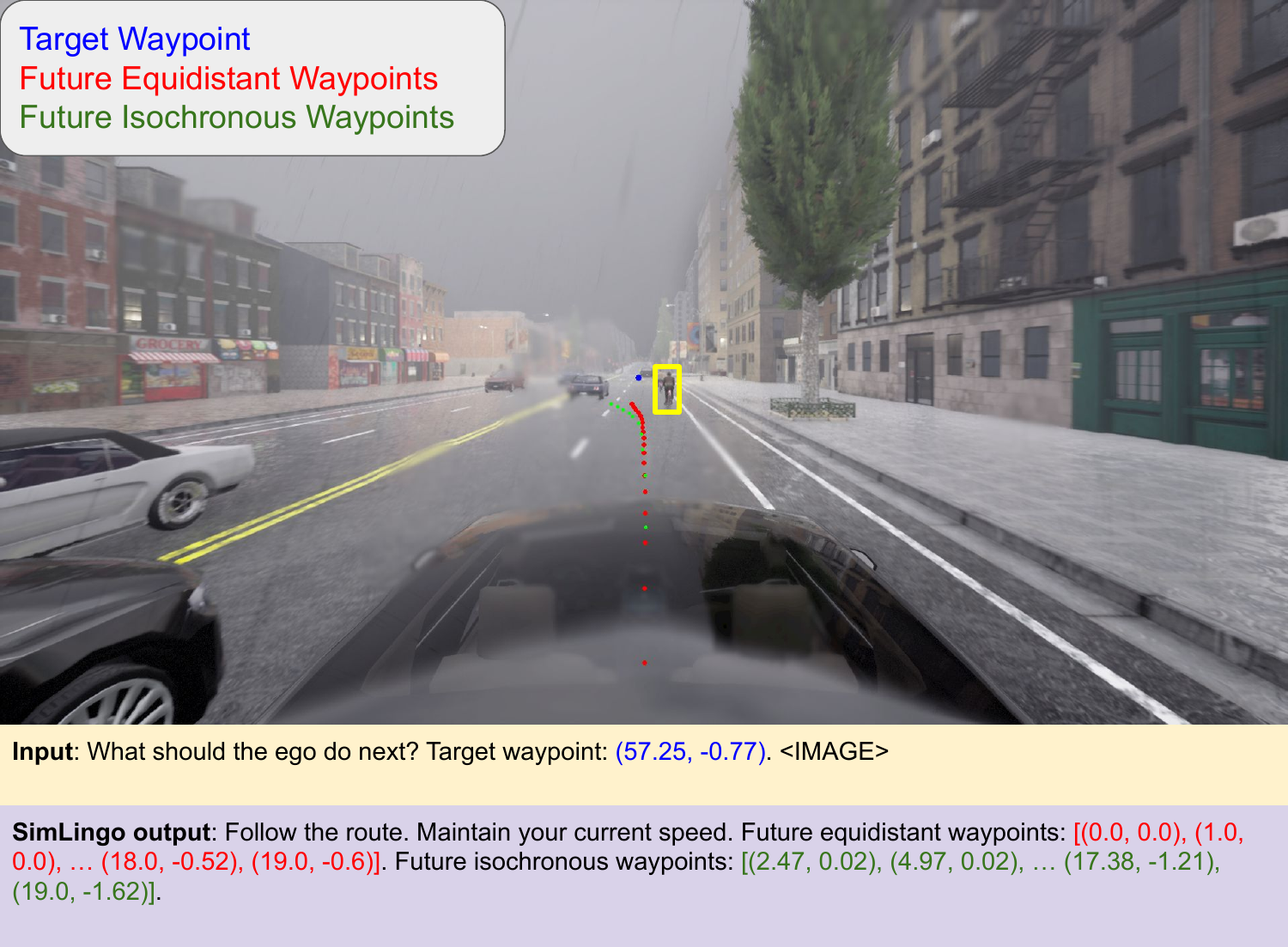}
            \caption{SimLingo~\cite{renz2025simlingo}
            }
            \label{fig:qualitative_simlingo_1}
        \end{subfigure}
        \hfill
        \begin{subfigure}[t]{0.33\textwidth}
            \centering 
            \includegraphics[width=0.99\textwidth]{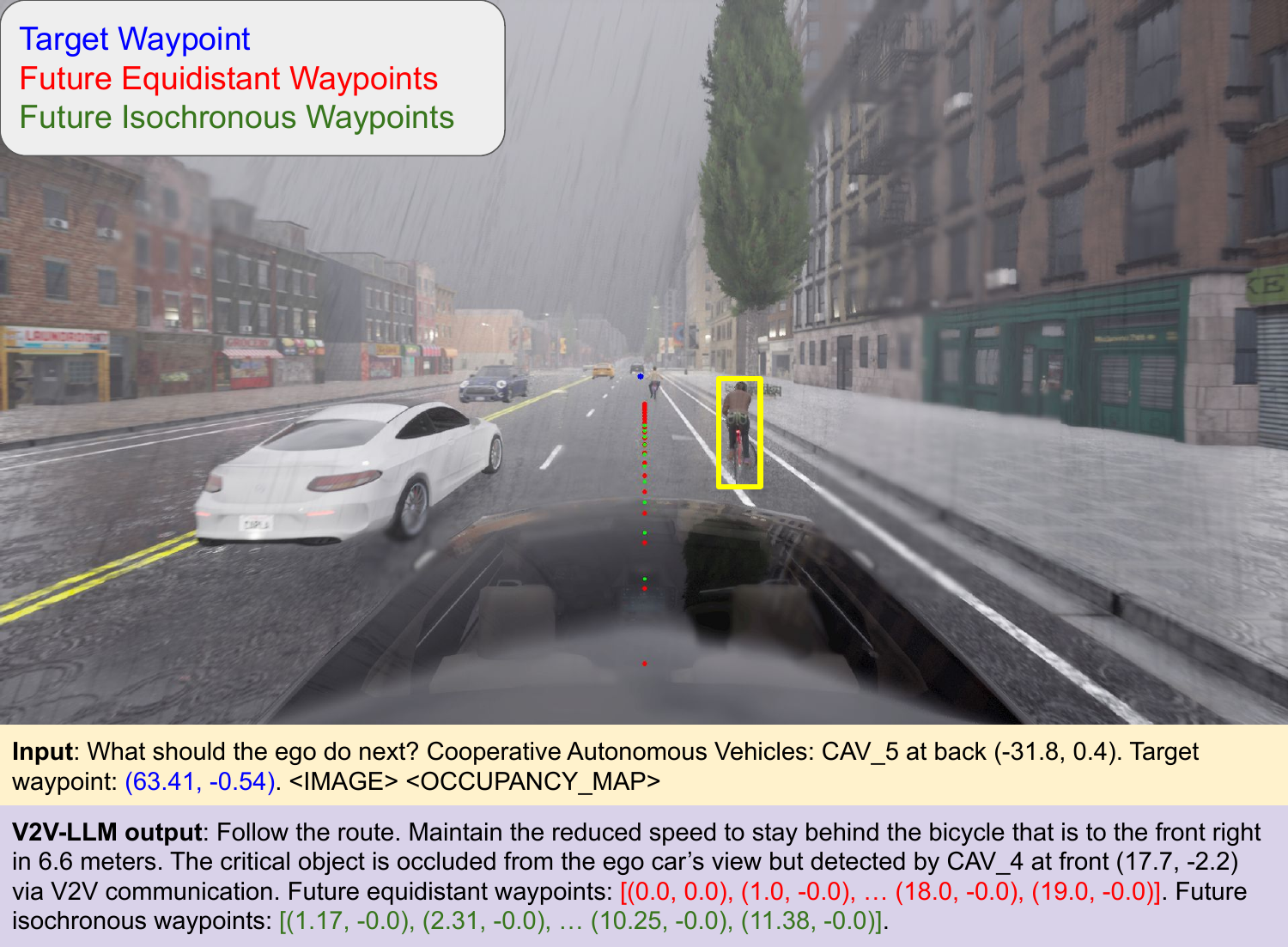}
            \caption{V2V-LLM~\cite{chiu2026v2vllm}
            }
            \label{fig:qualitative_v2vllm_1}
        \end{subfigure}
        
        \vspace{-10pt}
        \caption[]
        {
        Qualitative results in a rainy-day scenario with bicycles. 
        \textbf{Left} (\namemethod~(ours)): $CAV_0$ is able to reduce speed to stay behind the front bicycle when it is not clearly recognizable due to the rain-induced image blur, by utilizing the perception information shared from the front $CAV_4$.
        \textbf{Middle} (SimLingo~\cite{renz2025simlingo}): the language output does not mention the bicycle, while the action output indicates lane changing to avoid the potential collision with the bicycle. \textbf{Right} (V2V-LLM~\cite{chiu2026v2vllm}): the action output does not involve lane changing. 
        } 
        \label{fig:qualitative_1}
        \vspace{-10pt}
\end{figure*} 

\begin{figure*}[!t]
        \centering
        \begin{subfigure}[t]{0.33\textwidth}
            \centering 
            \includegraphics[width=0.99\textwidth]{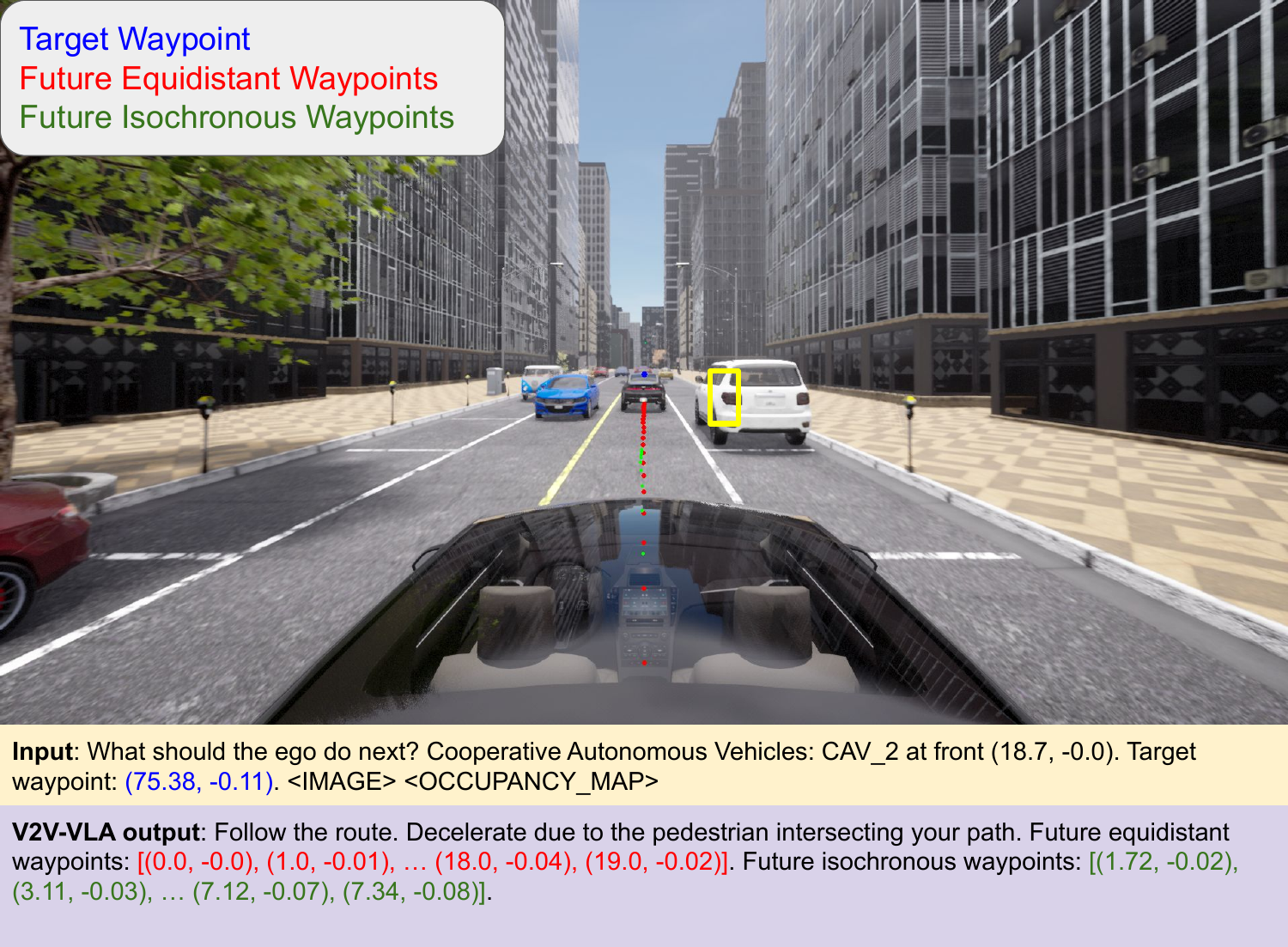}
            \caption{\namemethod~(ours)
            }
            \label{fig:qualitative_v2vvla_2}
        \end{subfigure}
        \hfill
        \begin{subfigure}[t]{0.33\textwidth}  
            \centering 
            \includegraphics[width=0.99\textwidth]{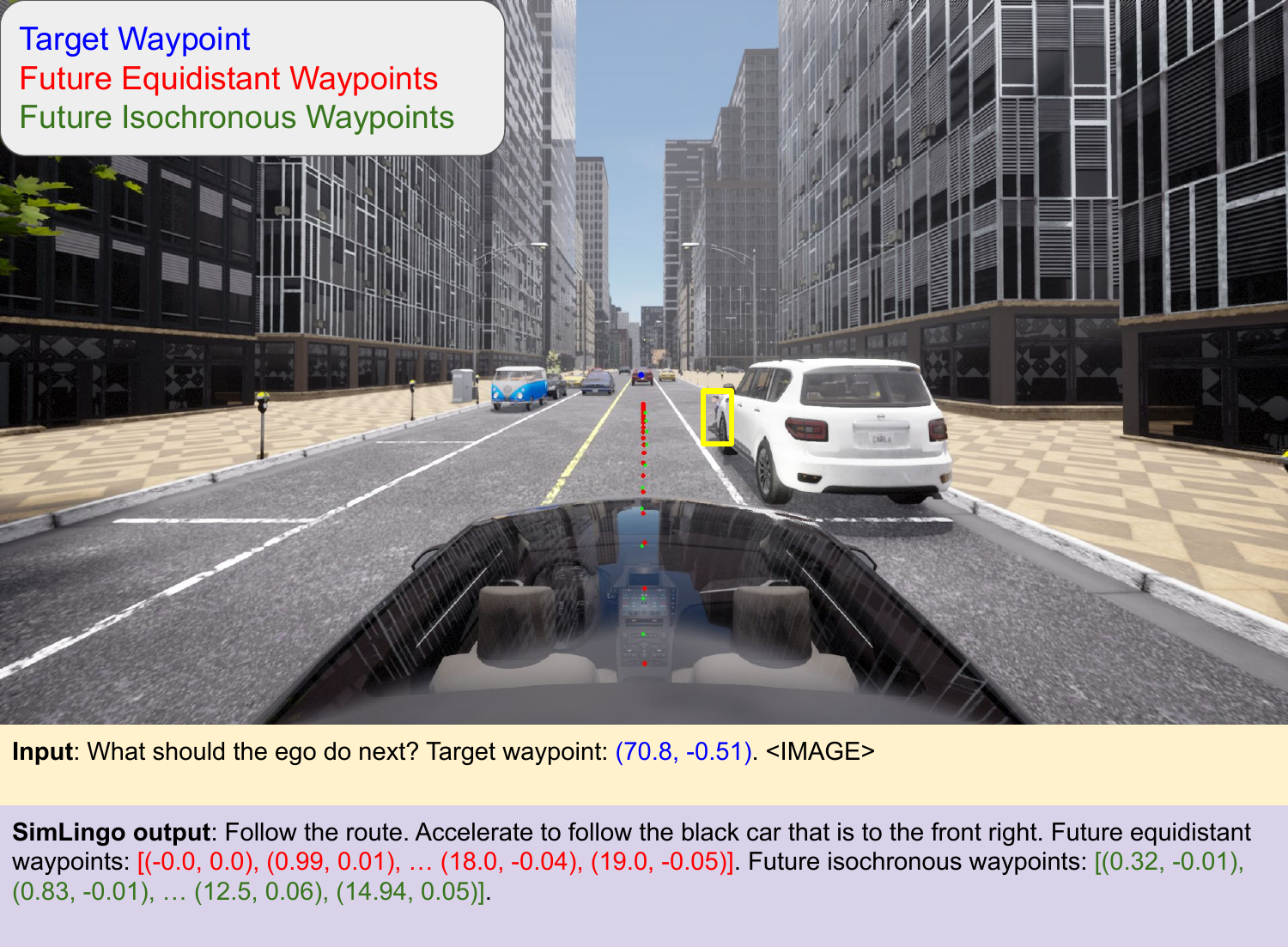}
            \caption{SimLingo~\cite{renz2025simlingo}
            }
            \label{fig:qualitative_simlingo_2}
        \end{subfigure}
        \hfill
        \begin{subfigure}[t]{0.33\textwidth}
            \centering 
            \includegraphics[width=0.99\textwidth]{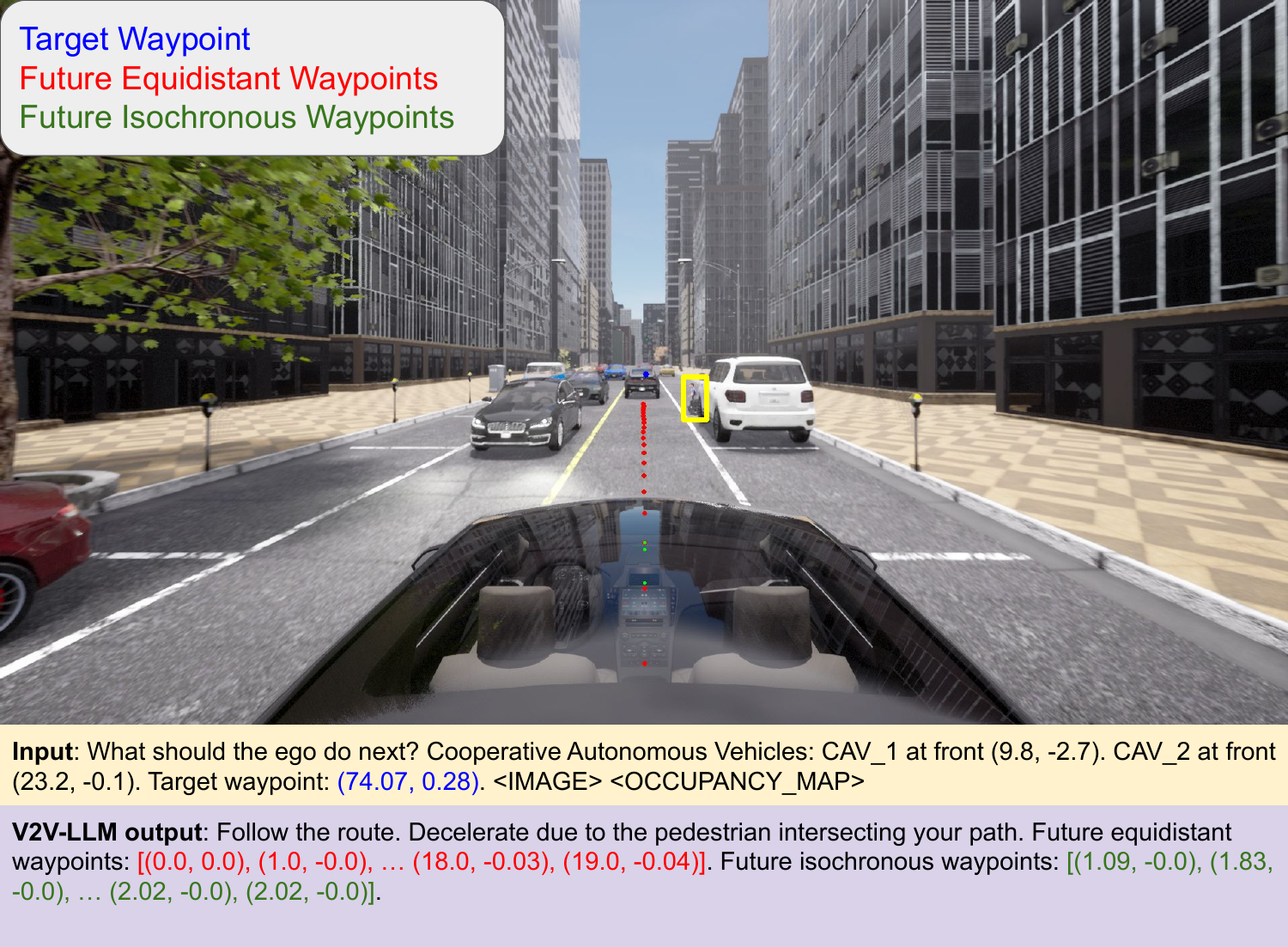}
            \caption{V2V-LLM~\cite{chiu2026v2vllm}
            }
            \label{fig:qualitative_v2vllm_2}
        \end{subfigure}
        
        \vspace{-10pt}
        \caption[]
        {
        Qualitative results in a sunny-day setting with a pedestrian crossing the street from an occluded region behind a parked vehicle. 
        \textbf{Left} (\namemethod~(ours)): $CAV_0$ is able to decelerate when the pedestrian, located in the yellow bounding box, is still occluded by the white parked vehicle, by utilizing the perception information shared from the front $CAV_2$.
        \textbf{Middle} (SimLingo~\cite{renz2025simlingo}): still decide to accelerate when the pedestrian is partially visible. \textbf{Right} (V2V-LLM~\cite{chiu2026v2vllm}): start to decelerate when the pedestrian becomes partially visible.
        } 
        \label{fig:qualitative_2}
        \vspace{-10pt}
\end{figure*} 

\begin{figure*}[!t]
        \centering
        \begin{subfigure}[t]{0.33\textwidth}
            \centering 
            \includegraphics[width=0.99\textwidth]{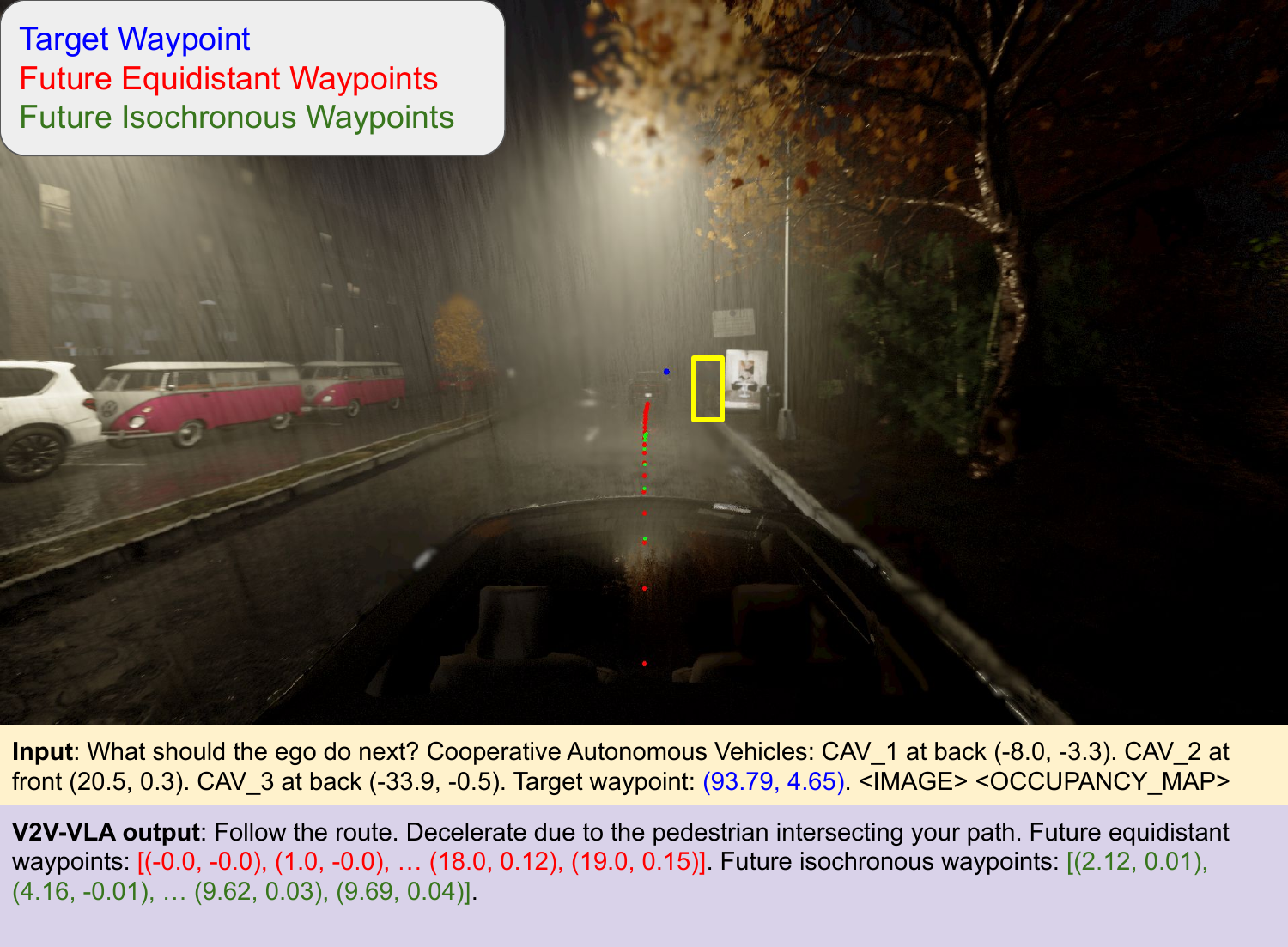}
            \caption{\namemethod~(ours)
            }
            \label{fig:qualitative_v2vvla_3}
        \end{subfigure}
        \hfill
        \begin{subfigure}[t]{0.33\textwidth}  
            \centering 
            \includegraphics[width=0.99\textwidth]{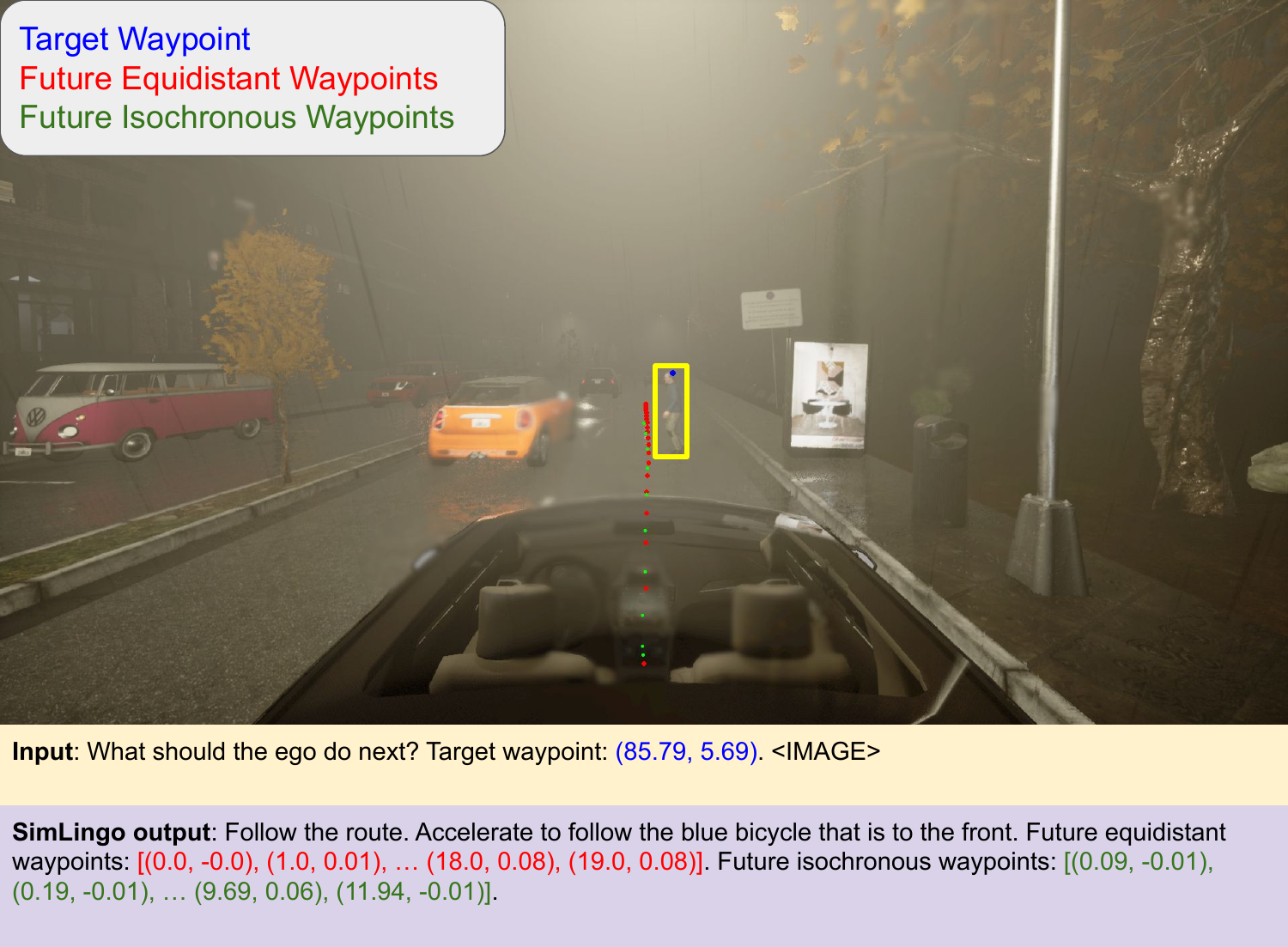}
            \caption{SimLingo~\cite{renz2025simlingo}
            }
            \label{fig:qualitative_simlingo_3}
        \end{subfigure}
        \hfill
        \begin{subfigure}[t]{0.33\textwidth}
            \centering 
            \includegraphics[width=0.99\textwidth]{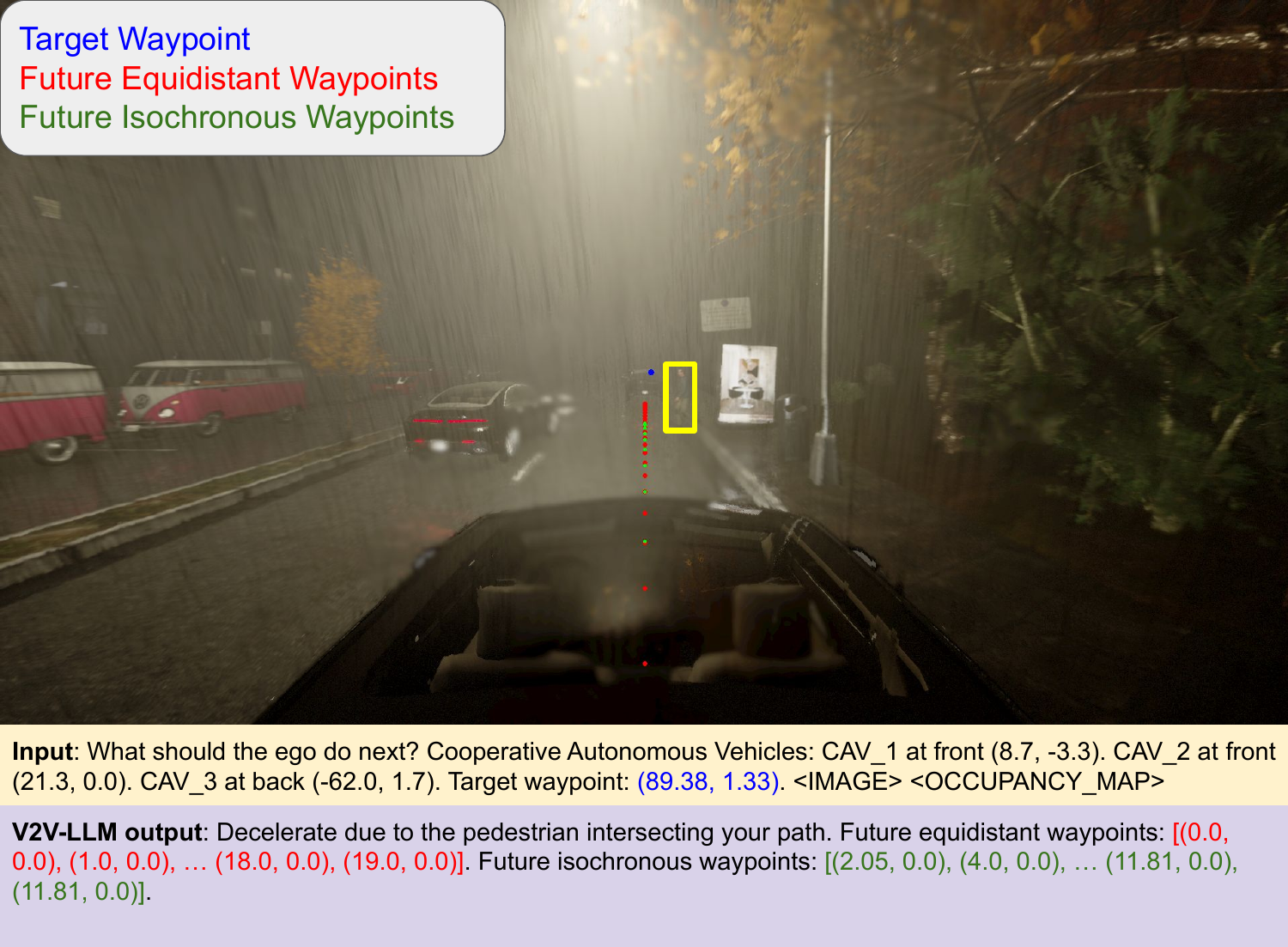}
            \caption{V2V-LLM~\cite{chiu2026v2vllm}
            }
            \label{fig:qualitative_v2vllm_3}
        \end{subfigure}
        
        \vspace{-10pt}
        \caption[]
        {
        Qualitative experimental result in a rainy-night setting with a pedestrian crossing the street.
        \textbf{Left} (\namemethod~(ours)): $CAV_0$ is able to decelerate sufficiently earlier when the pedestrian, located in the yellow bounding box, is still not clearly recognizable due to the lighting condition, by utilizing the perception information shared from the front $CAV_2$ and avoid the potential collision.
        \textbf{Middle} (SimLingo~\cite{renz2025simlingo}): the language output misclassifies the pedestrian as a bicycle and decide to accelerate when the pedestrian is recognizable. \textbf{Right} (V2V-LLM~\cite{chiu2026v2vllm}): start to decelerate when the pedestrian becomes partially recognizable.
        } 
        \label{fig:qualitative_3}
        \vspace{-15pt}
\end{figure*} 

\subsection{Qualitative Result}

\cref{fig:qualitative_1,fig:qualitative_2,fig:qualitative_3} show the qualitative results of \namemethod~in \namedataset, compared to SimLingo~\cite{renz2025simlingo} and V2V-LLM~\cite{chiu2026v2vllm}. In general, \namemethod~is able to utilize the perception information shared from a selected communicative vehicle and take appropriate actions earlier to avoid collisions with critical objects when the objects are undetectable or unrecognizable in the camera image of the ego vehicle due to occlusion or special weather and lighting conditions. More qualitative results and closed-loop evaluation videos can be found in the supplementary material.

\section{Conclusion}
\label{sec:conclusion}

In this paper, we have presented \namedataset, a benchmark for cooperative multi-agent, closed-loop, end-to-end driving with reasoning. Compared to prior related works, the proposed benchmark supports closed-loop simulation with a larger number of connected autonomous vehicles and various background traffic participants across more complex, safety-critical driving scenarios.
We have also introduced \namemethod, a novel vehicle-to-vehicle vision-language-action model for cooperative autonomous driving. \namemethod~aggregates information from a selected communicative vehicle and generates driving actions, reasoning, communication policies, and future waypoints. Our experimental results demonstrate that \namemethod~achieves better driving performance than state-of-the-art single-agent and multi-agent baselines. 
Our work provides a foundation for future research on multi-agent, closed-loop, end-to-end cooperative autonomous driving with reasoning.
To further facilitate open-source research, we will publicly release our code, benchmark, and model checkpoints.
\section*{LLM Usage Statement}
ChatGPT is used to fix grammar errors in writing and draw the glowing curves and circles on top of our CARLA rendered image in \cref{fig:teaser}.
{
    \small
    \bibliographystyle{ieeenat_fullname}
    \bibliography{main}
}

\clearpage
\setcounter{page}{1}
\maketitlesupplementary

\section{\namedataset~Statistics}
\begin{figure}[tb]
  \centering
  \includegraphics[width=0.45\textwidth]{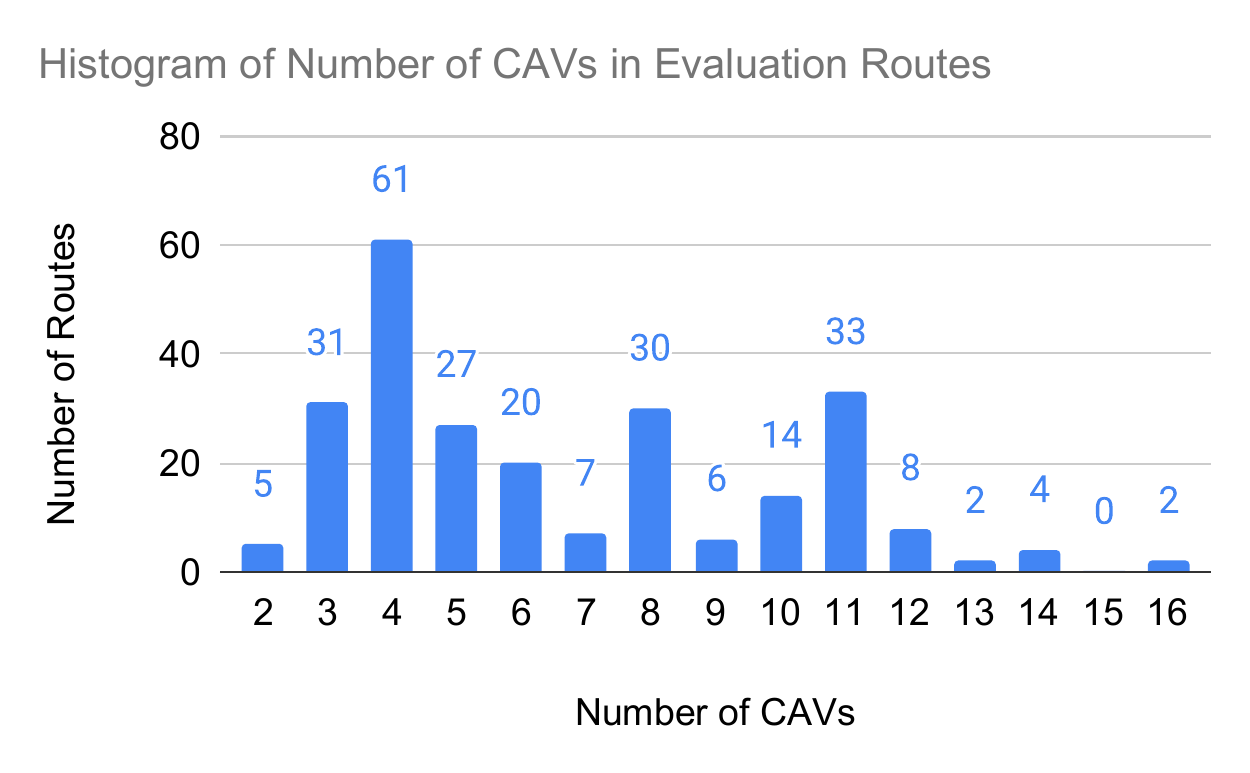}
  \vspace{-15pt}
  \caption{Histogram of the number of connected autonomous vehicles (CAVs) in the evaluation routes of \namedataset.
  }
  \label{fig:histogram}
  \vspace{-15pt}
\end{figure}

\namedataset~benchmark includes a total of $50$ types of scenario, and each scenario is instantiated under $5$ weather and lighting conditions, resulting in a total of $250$ evaluation routes. The $50$ scenarios include $44$ scenarios expanded from Bench2Drive~\cite{jia2024bench}'s single-agent setting to our multi-agent setting with $2$ to $16$ connected autonomous vehicles (CAVs). In addition to the scenarios expanded from~\cite{jia2024bench}, \namedataset~benchmark also includes $6$ more scenarios with large vehicles, such as busses and trucks: \textit{LargeParkingCrossingPedestrian}, \textit{LargeParkingCutIn}, \textit{LargeSignalizedJunctionLeftTurn}, \textit{LargeStaticCutIn}, \textit{LargeVehicleTurningRoute}, and \textit{LargeVehicleTurningRoutePedestrian}, resulting in $30$ more evaluation routes. Detailed description of each of the $50$ scenarios types can be seen in \cref{tab:scenraio_description_1,tab:scenraio_description_2}.
\cref{fig:histogram} shows the histogram of the number of connected autonomous vehicles (CAVs) on the evaluation routes of \namedataset.

\section{Driving Scenarios and Skill Categorization}
\label{sec:skill_categorization}
\begin{figure}[tb]
  \centering
  \includegraphics[width=0.45\textwidth]{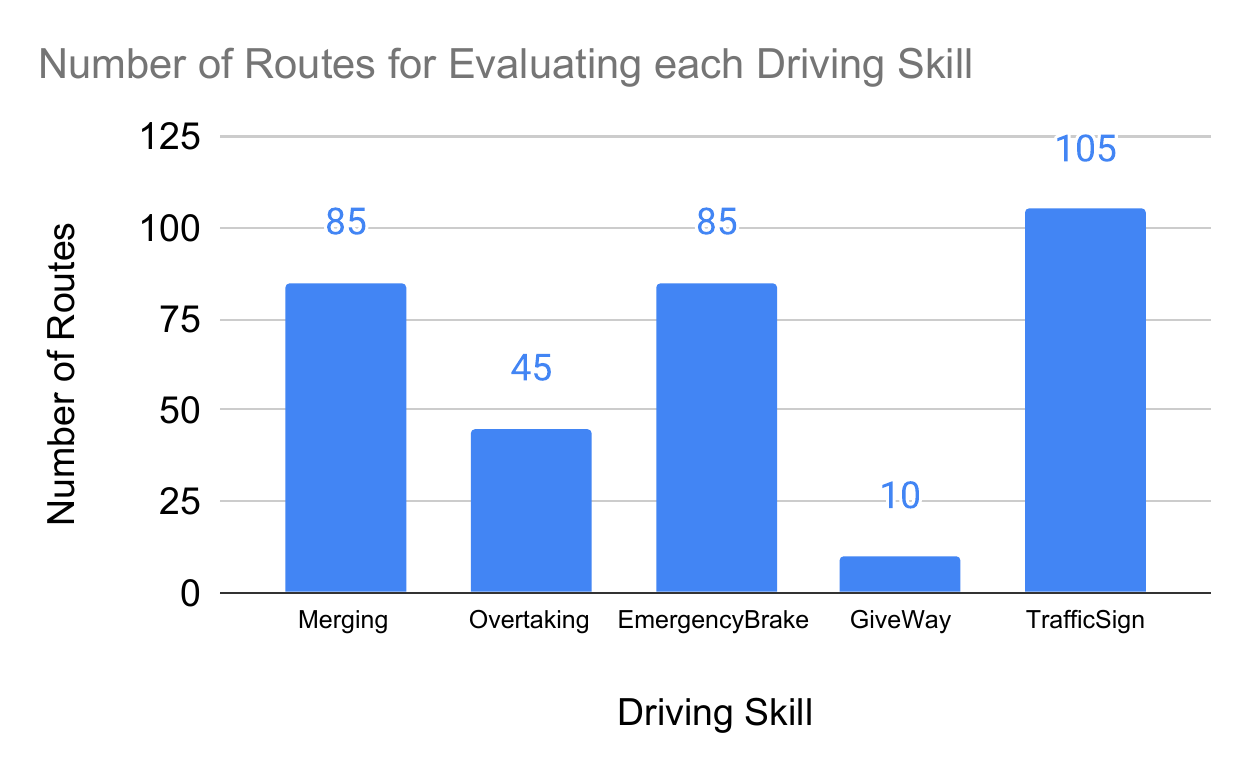}
  \vspace{-15pt}
  \caption{Distribution of the evaluation routes evaluating each driving skill in \namedataset.
  }
  \label{fig:skill}
  \vspace{-10pt}
\end{figure}

Each scenario in \namedataset~is designed to evaluate one or more specific driving skills, including \textit{Merging}, \textit{Overtaking}, \textit{EmergencyBrake}, \textit{GiveWay}, and \textit{TrafficSign} understanding. \cref{fig:skill} shows the distribution of \namedataset's evaluation routes that evaluate each driving skill. The detailed skill categorization is summarized in 
\cref{tab:evaluation_route_skill}.

\begin{table}[tb]
  \caption{Infraction score penalty factors.
  }
  \vspace{-10pt}
  \label{tab:infraction_score}
  \setlength{\tabcolsep}{6pt}
  \centering
  \begin{tabular}{l | c }
    \toprule
    Infraction & Penalty Factor \\
    \midrule
    Collisions with pedestrians & 0.50 \\
    Collisions with other vehicles & 0.60 \\
    Collisions with static elements & 0.65 \\
    Running a red light & 0.70 \\
    Failure to yield to emergency vehicle & 0.70 \\
    Running a stop sign & 0.80 \\
    Scenario timeout & 0.70 \\
  \bottomrule
  \end{tabular}
  \vspace{-10pt}
\end{table}

\section{Infraction Score Penalty Factors}
\cref{tab:infraction_score} summarizes the penalty factors for infraction scores for each type of driving failure. More details of each infraction can be seen in the CARLA~\cite{dosovitskiy2017carla} autonomous driving challenge.

\section{Visualization of BEV Occupancy Maps}

\begin{figure*}[t]
    \centering
    \begin{subfigure}{0.32\textwidth}
        \centering
        \includegraphics[width=\linewidth]{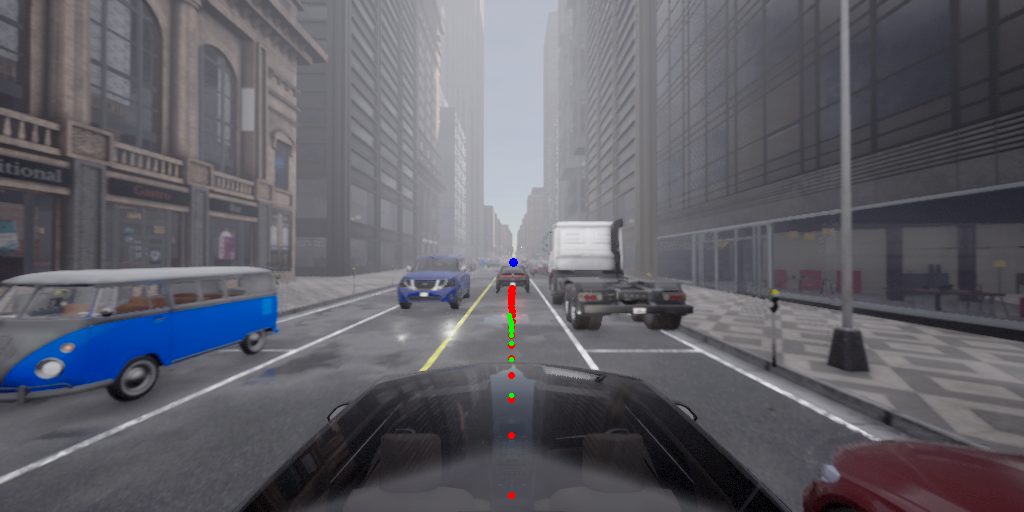}
        \caption{Frame $t$}
    \end{subfigure}
    \hfill
    \begin{subfigure}{0.32\textwidth}
        \centering
        \includegraphics[width=\linewidth]{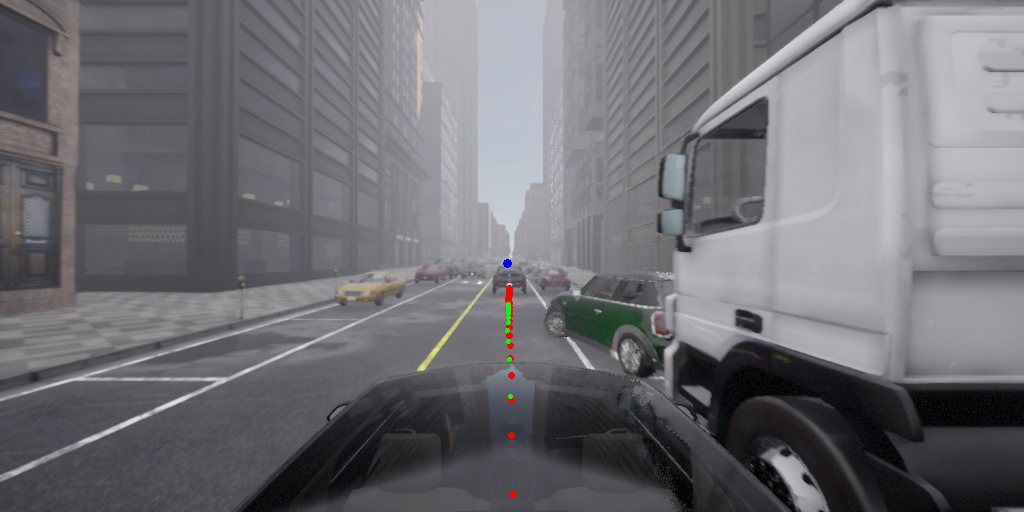}
        \caption{Frame $t + 1.5$ seconds}
    \end{subfigure}
    \hfill
    \begin{subfigure}{0.32\textwidth}
        \centering
        \includegraphics[width=\linewidth]{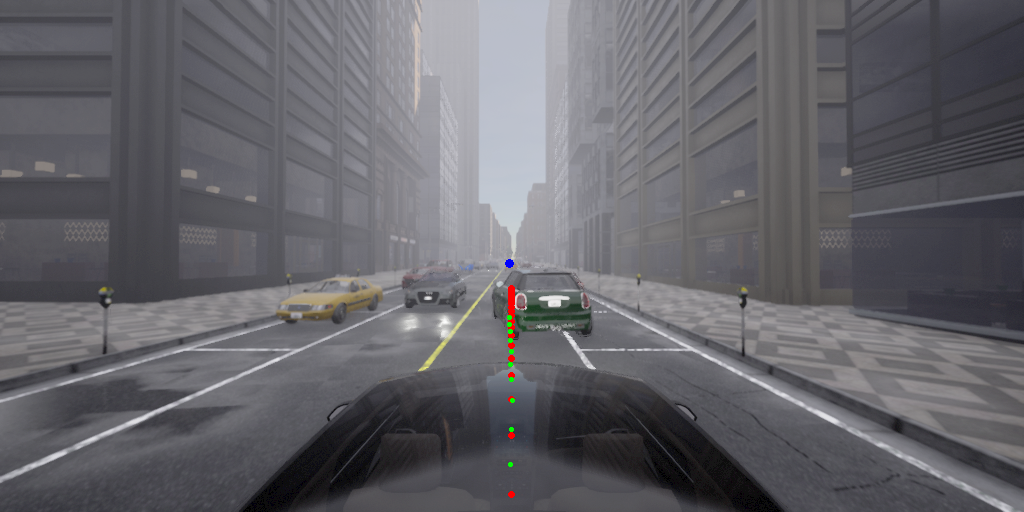}
        \caption{Frame $t + 3$ seconds}
    \end{subfigure}

    \vspace{-10pt}
    \caption{In this evaluation route, there is a green vehicle occluded by the front right white parked truck at frame $t$. That occluded green vehicle is cutting into the driving lane. The ego car $CAV_{ego}$ utilizes the perception information from its selected communicative vehicle $CAV_{c}$ in front and decelerates before the green conflicting vehicle becomes visible to the ego car $CAV_{ego}$ to avoid the potential collision.}
    \label{fig:example}
\end{figure*}

In this section, we show the  visualization of Bird-Eye-View (BEV) occupancy maps used in \namemethod's second vision input, and the relevant camera images in an example frame of an evaluation route in \namedataset, as shown in \cref{fig:example}.

\begin{figure*}[t]
    \centering
    \begin{subfigure}{0.32\textwidth}
        \centering
        \includegraphics[width=\linewidth]{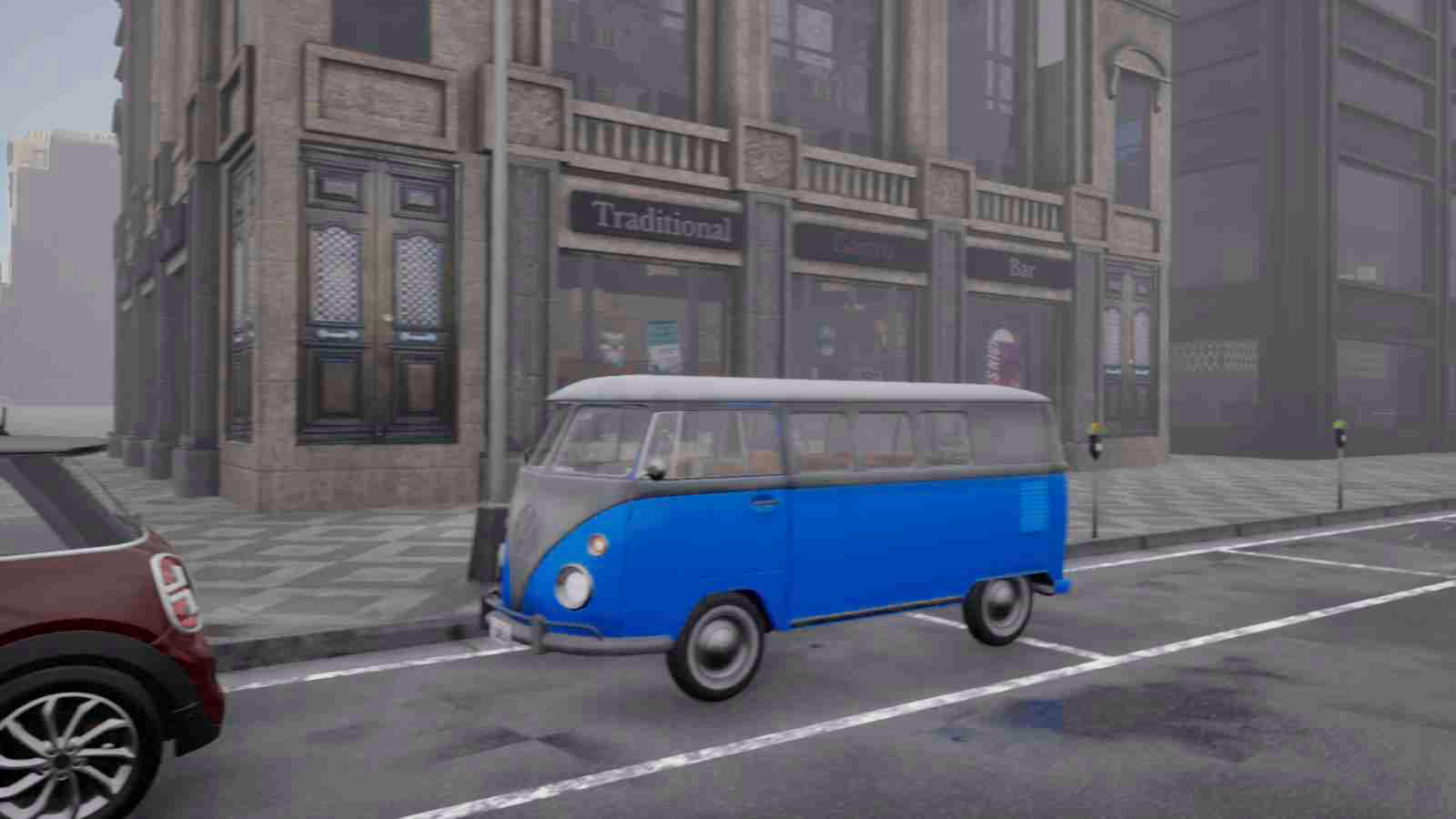}
        \caption{Front left}
    \end{subfigure}
    \hfill
    \begin{subfigure}{0.32\textwidth}
        \centering
        \includegraphics[width=\linewidth]{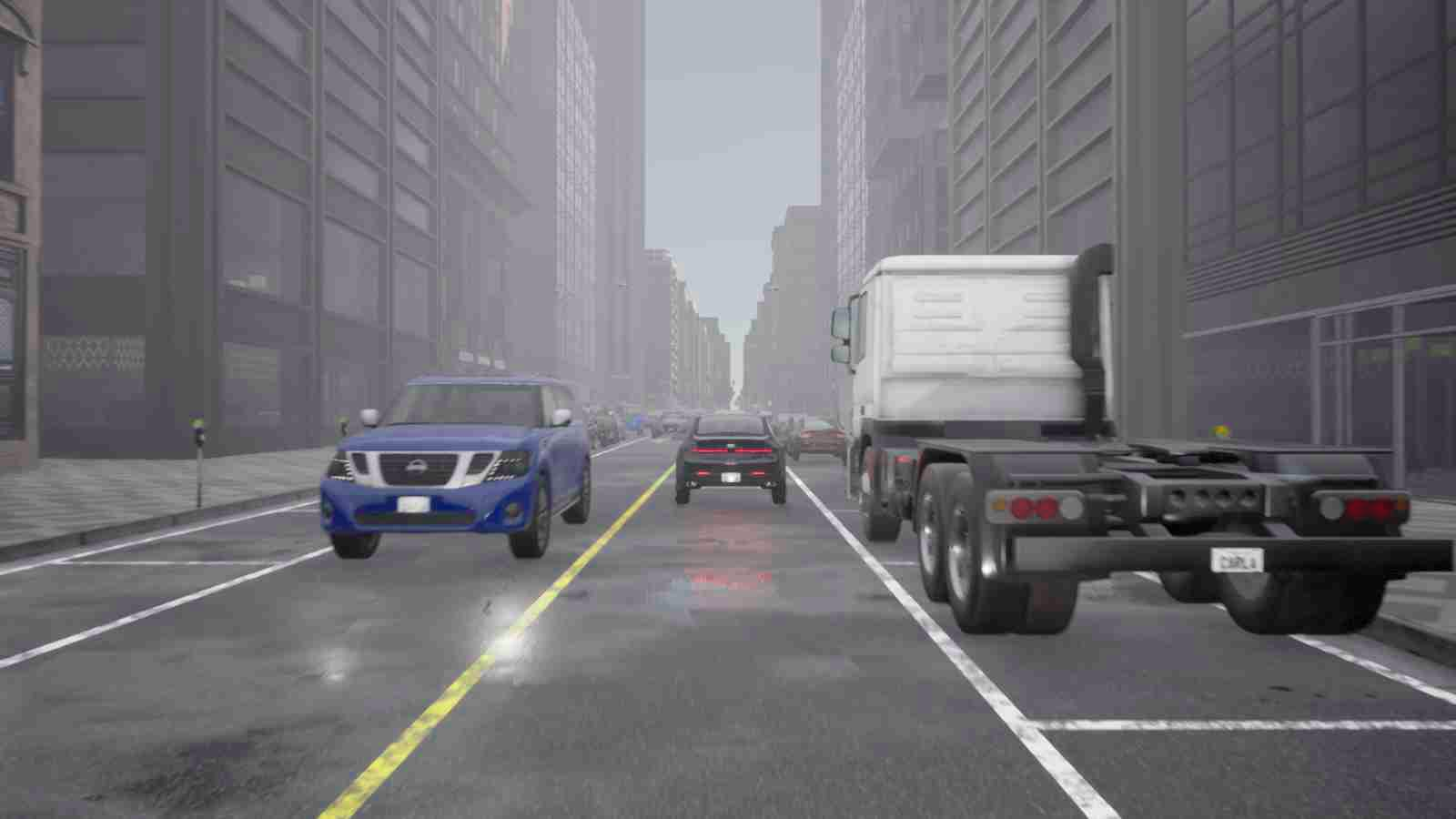}
        \caption{Front}
    \end{subfigure}
    \hfill
    \begin{subfigure}{0.32\textwidth}
        \centering
        \includegraphics[width=\linewidth]{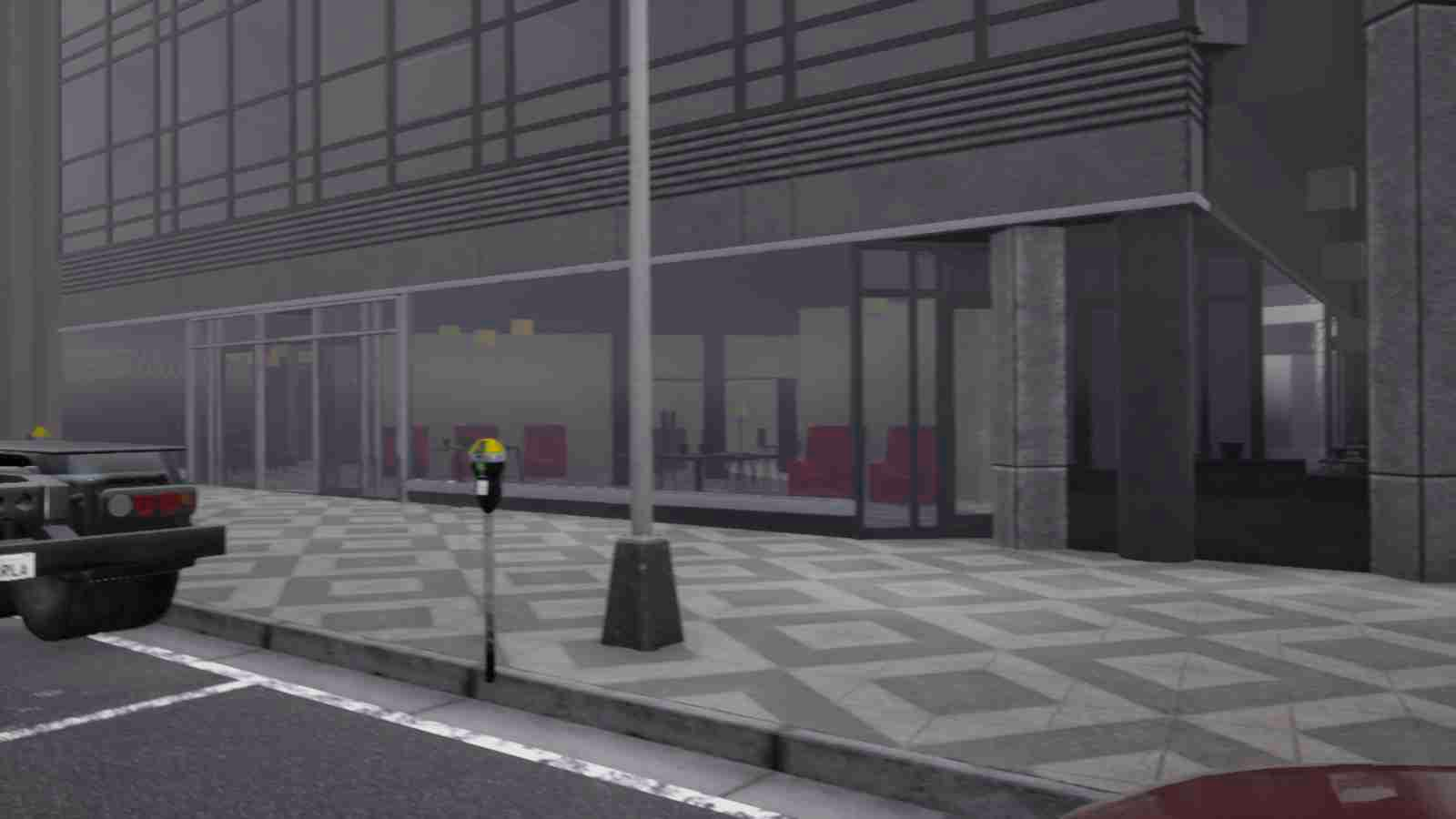}
        \caption{Front right}
    \end{subfigure}


    \begin{subfigure}{0.32\textwidth}
        \centering
        \includegraphics[width=\linewidth]{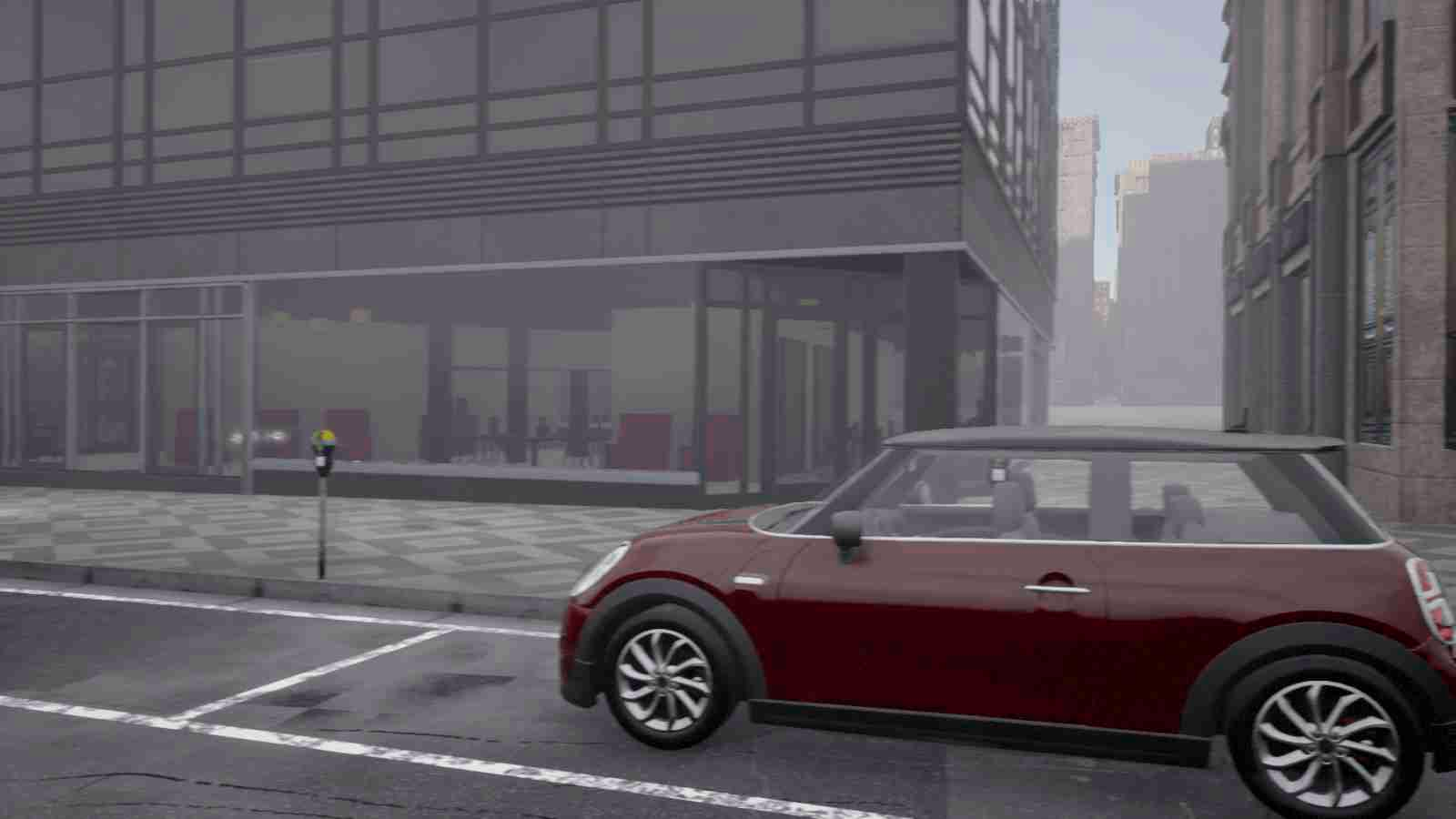}
        \caption{Back left}
    \end{subfigure}
    \hfill
    \begin{subfigure}{0.32\textwidth}
        \centering
        \includegraphics[width=\linewidth]{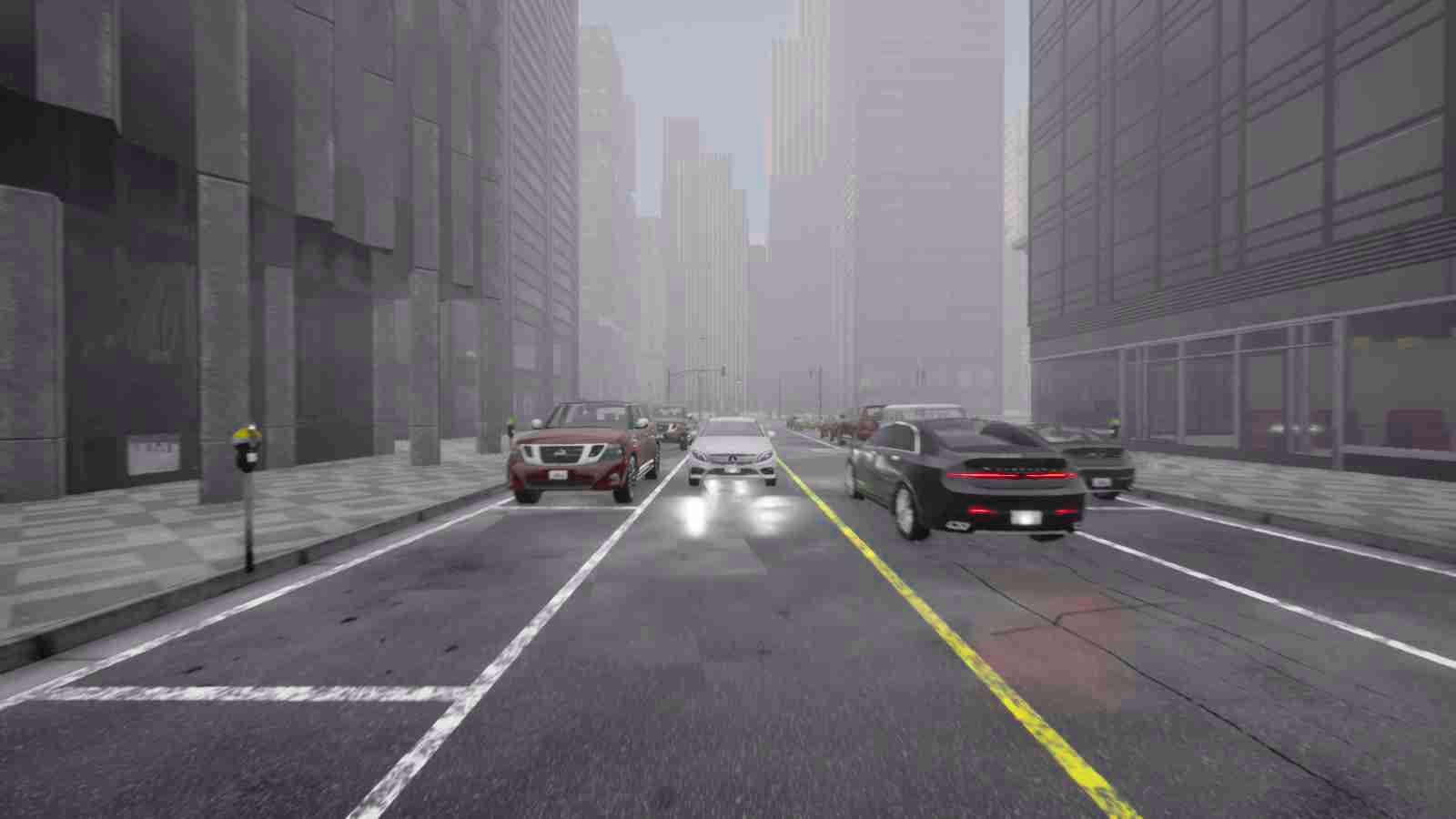}
        \caption{Back}
    \end{subfigure}
    \hfill
    \begin{subfigure}{0.32\textwidth}
        \centering
        \includegraphics[width=\linewidth]{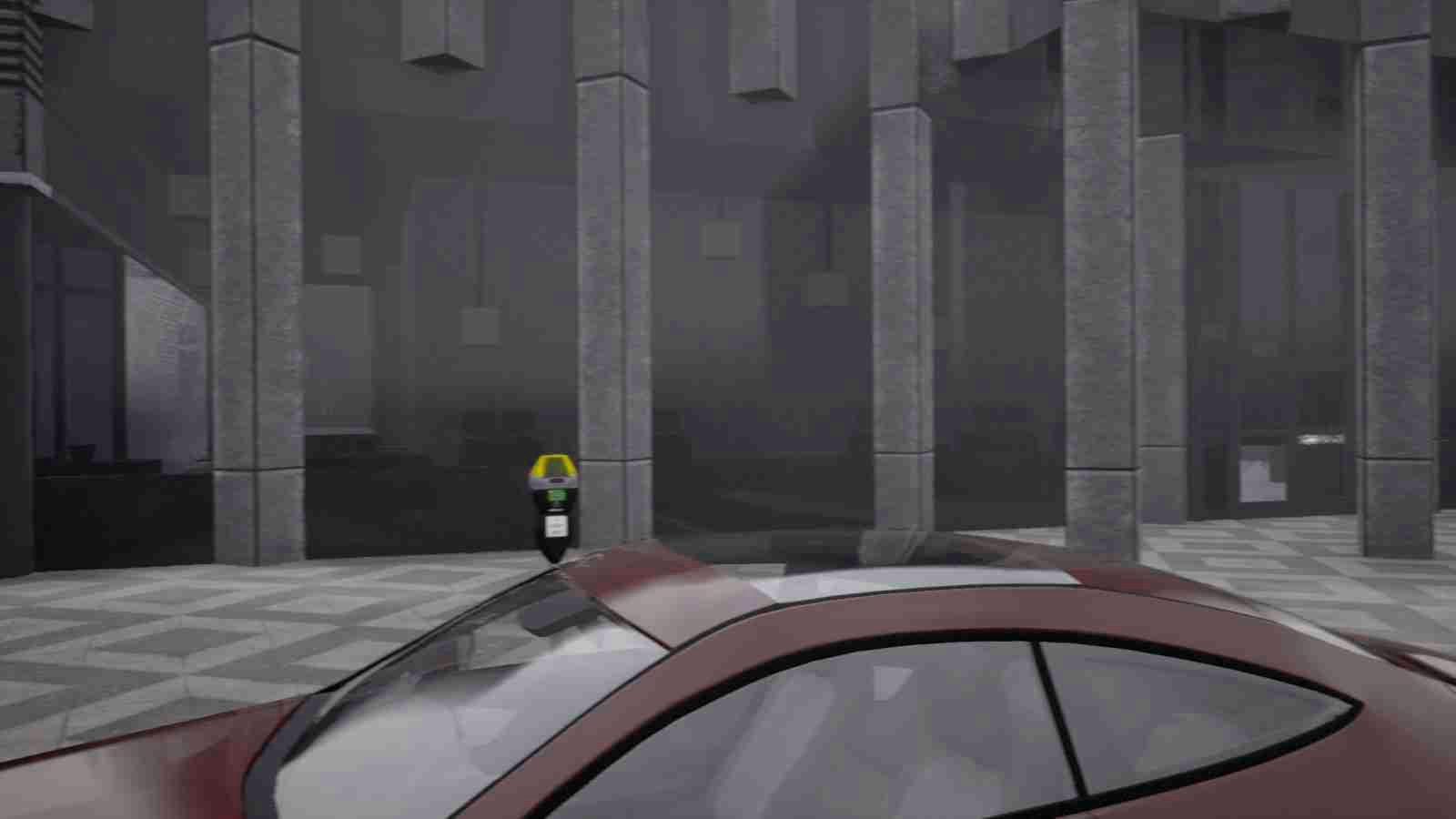}
        \caption{Back right}
    \end{subfigure}

    \vspace{-10pt}
    \caption{Six camera images of $CAV_{ego}$ at frame $t$ that generate the BEV occupancy map $O_{ego}$.}
    \label{fig:ego_six_camera_images}
\end{figure*}

\begin{figure*}[t]
    \centering
    \begin{subfigure}{0.32\textwidth}
        \centering
        \includegraphics[width=\linewidth]{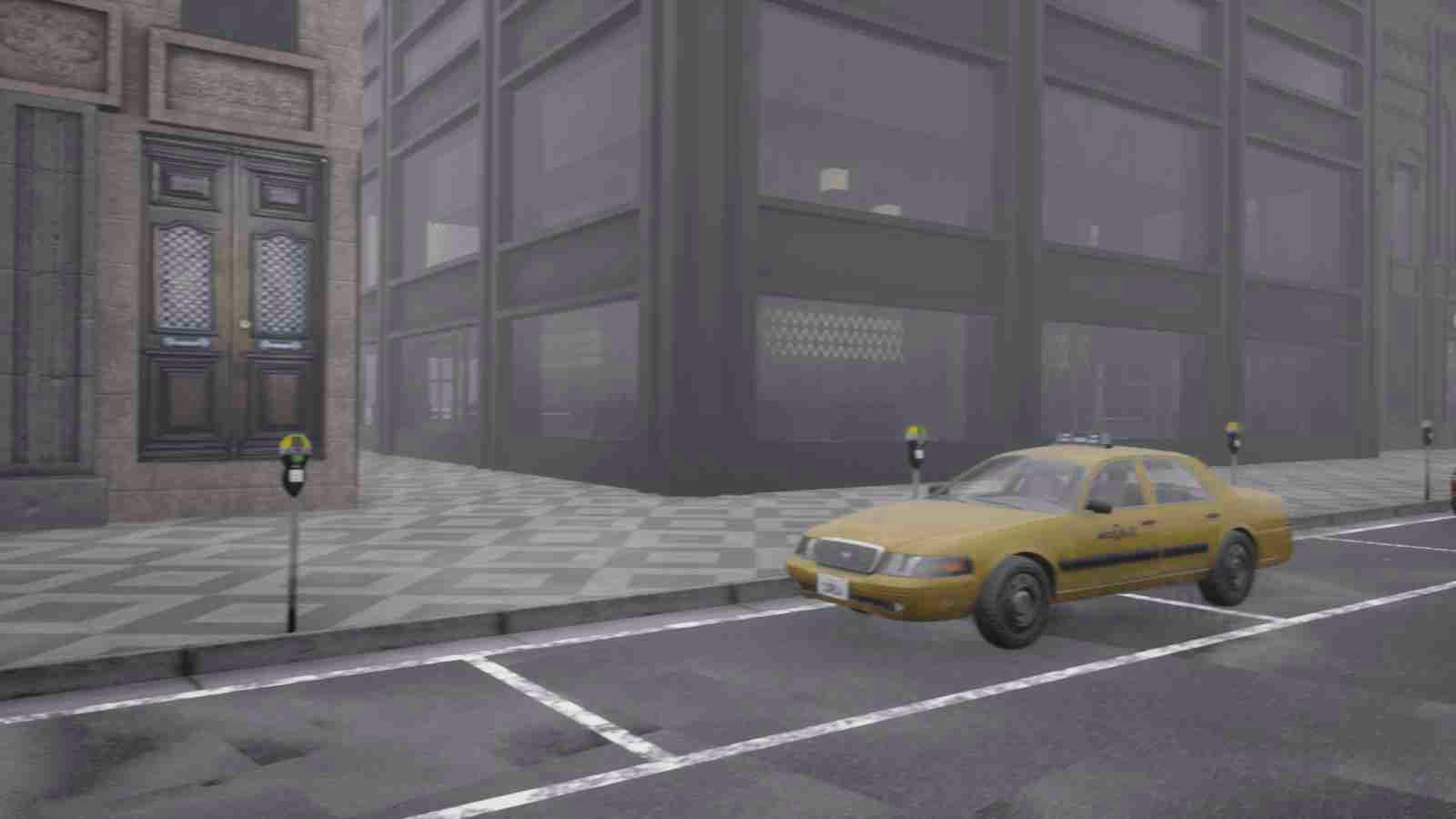}
        \caption{Front left}
    \end{subfigure}
    \hfill
    \begin{subfigure}{0.32\textwidth}
        \centering
        \includegraphics[width=\linewidth]{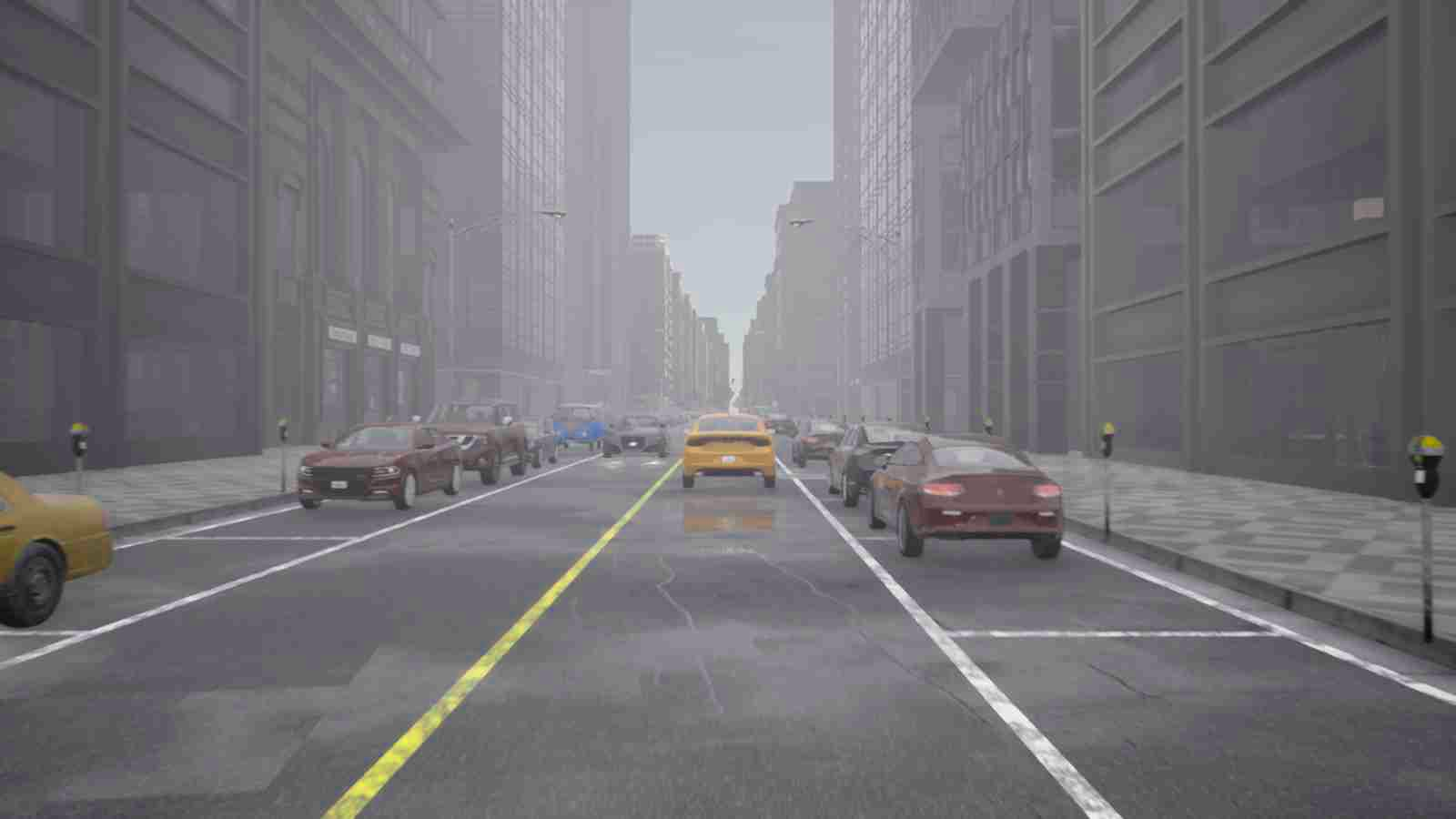}
        \caption{Front}
    \end{subfigure}
    \hfill
    \begin{subfigure}{0.32\textwidth}
        \centering
        \includegraphics[width=\linewidth]{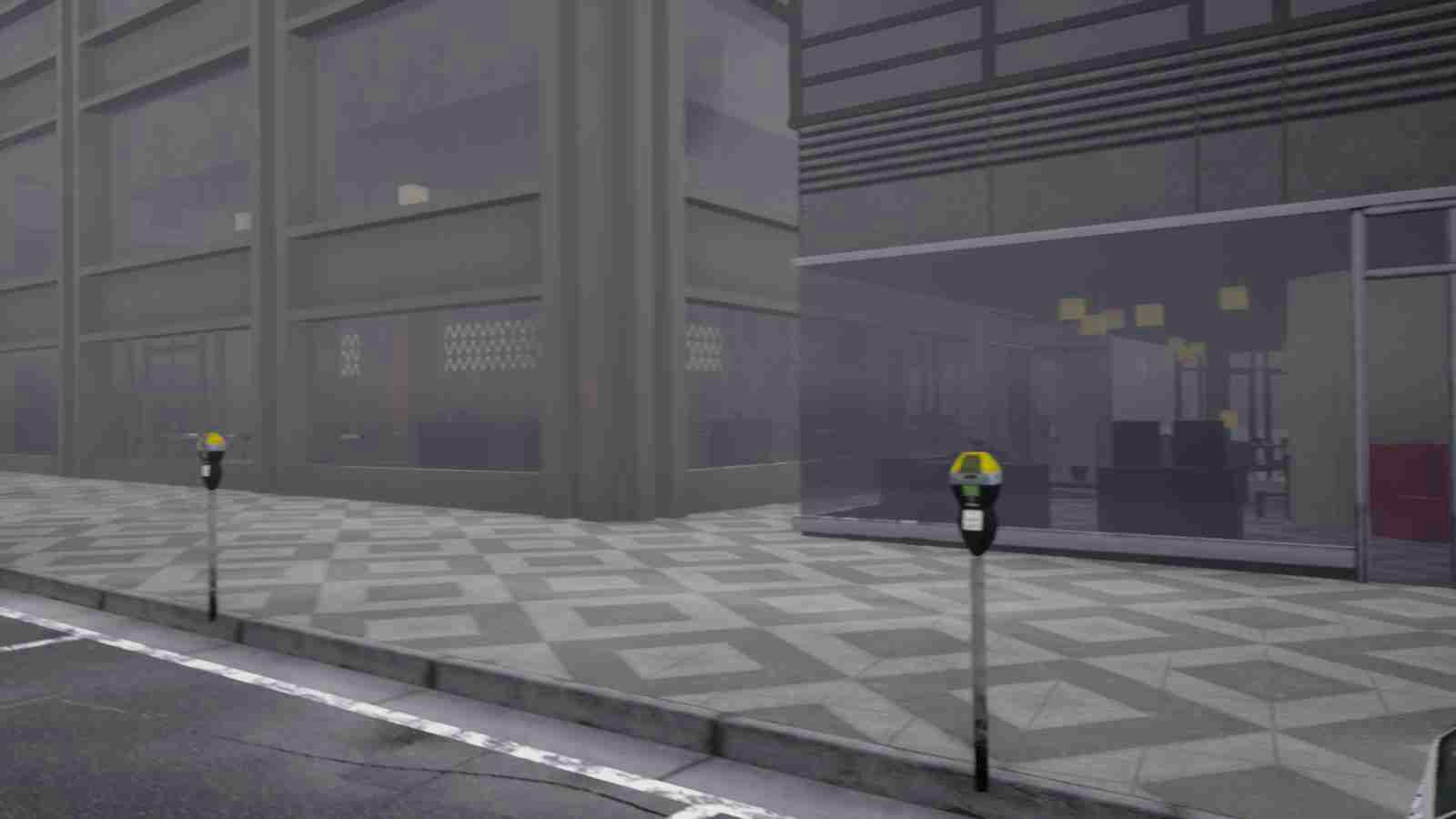}
        \caption{Front right}
    \end{subfigure}


    \begin{subfigure}{0.32\textwidth}
        \centering
        \includegraphics[width=\linewidth]{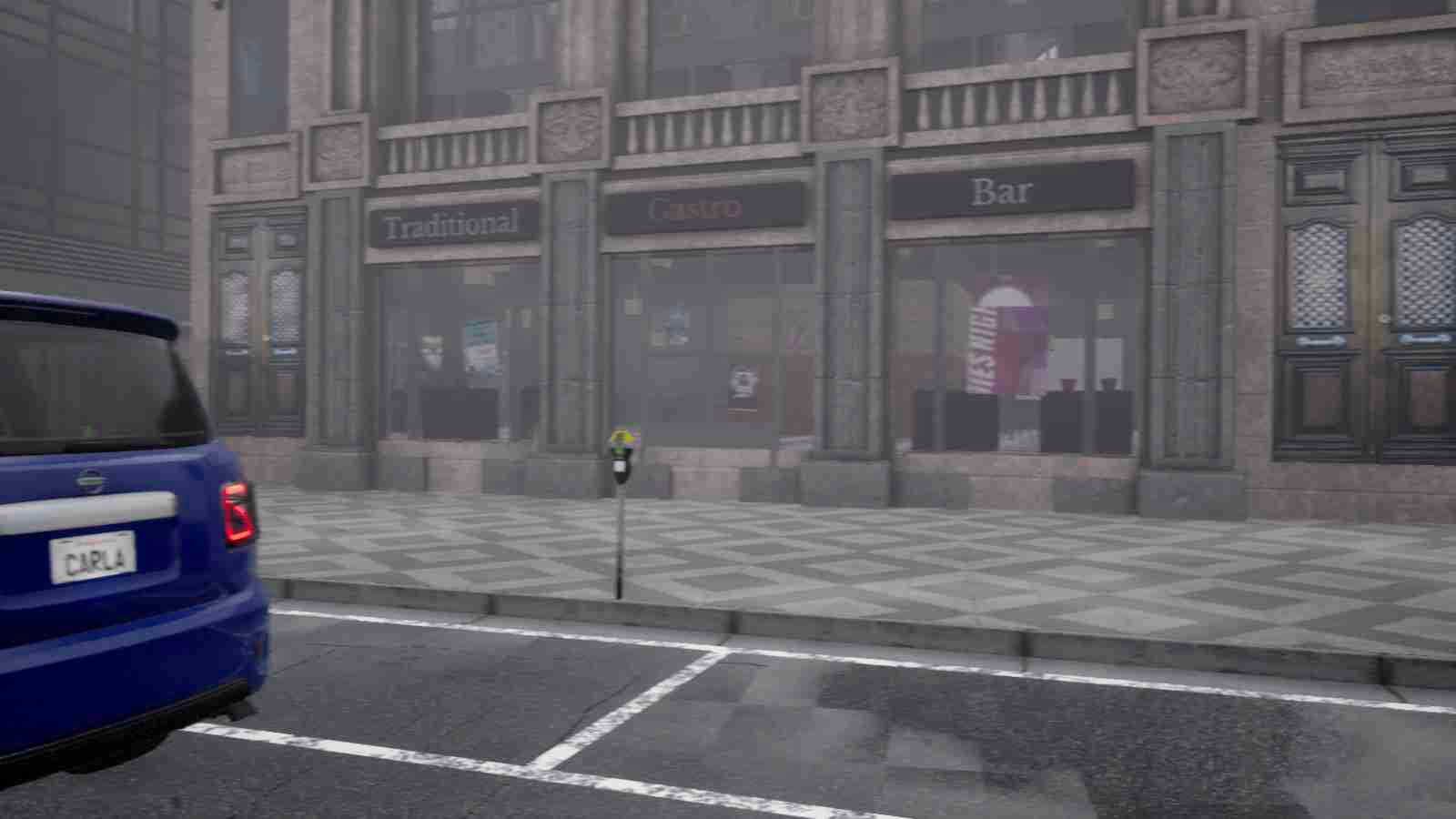}
        \caption{Back left}
    \end{subfigure}
    \hfill
    \begin{subfigure}{0.32\textwidth}
        \centering
        \includegraphics[width=\linewidth]{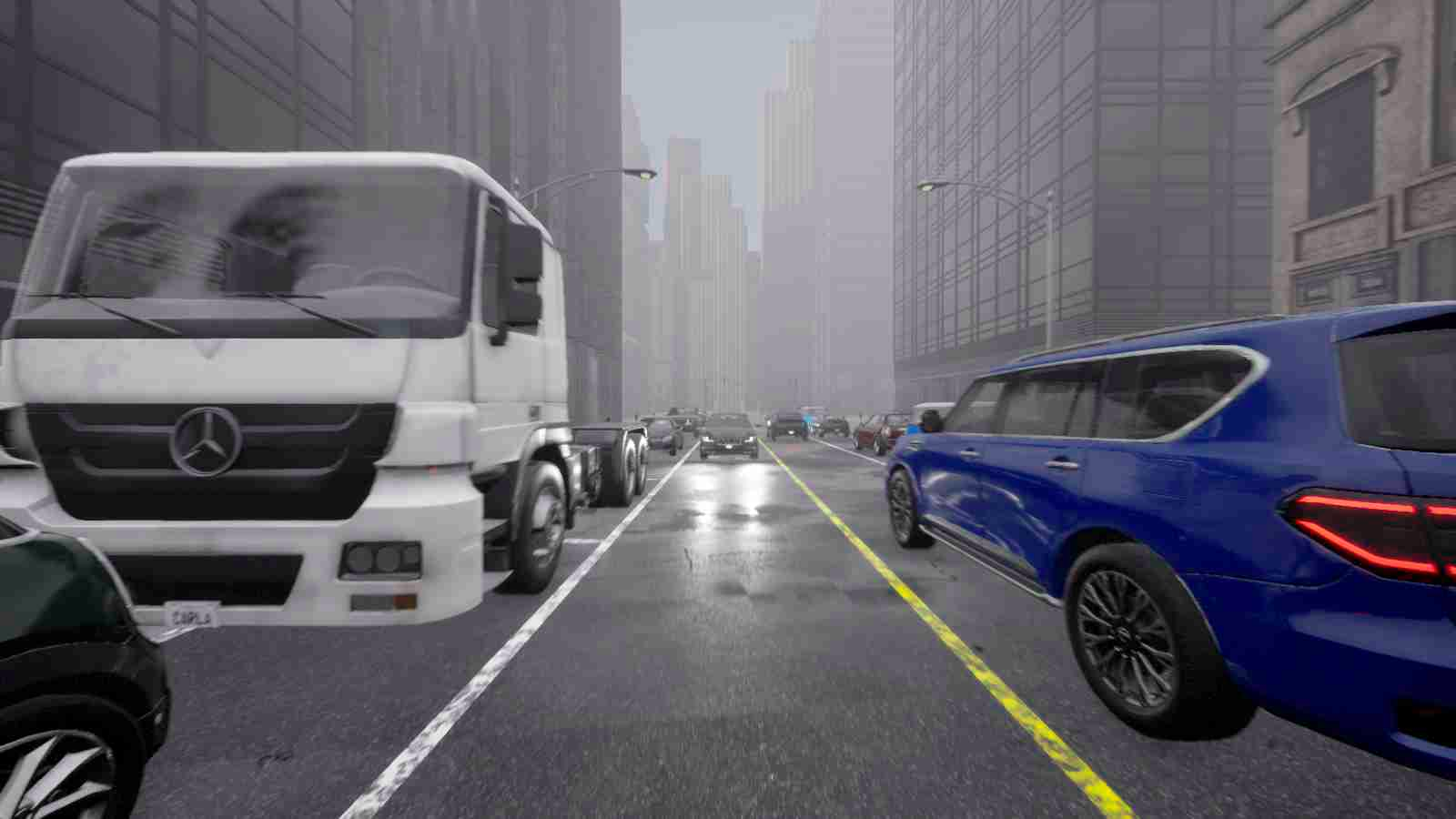}
        \caption{Back}
    \end{subfigure}
    \hfill
    \begin{subfigure}{0.32\textwidth}
        \centering
        \includegraphics[width=\linewidth]{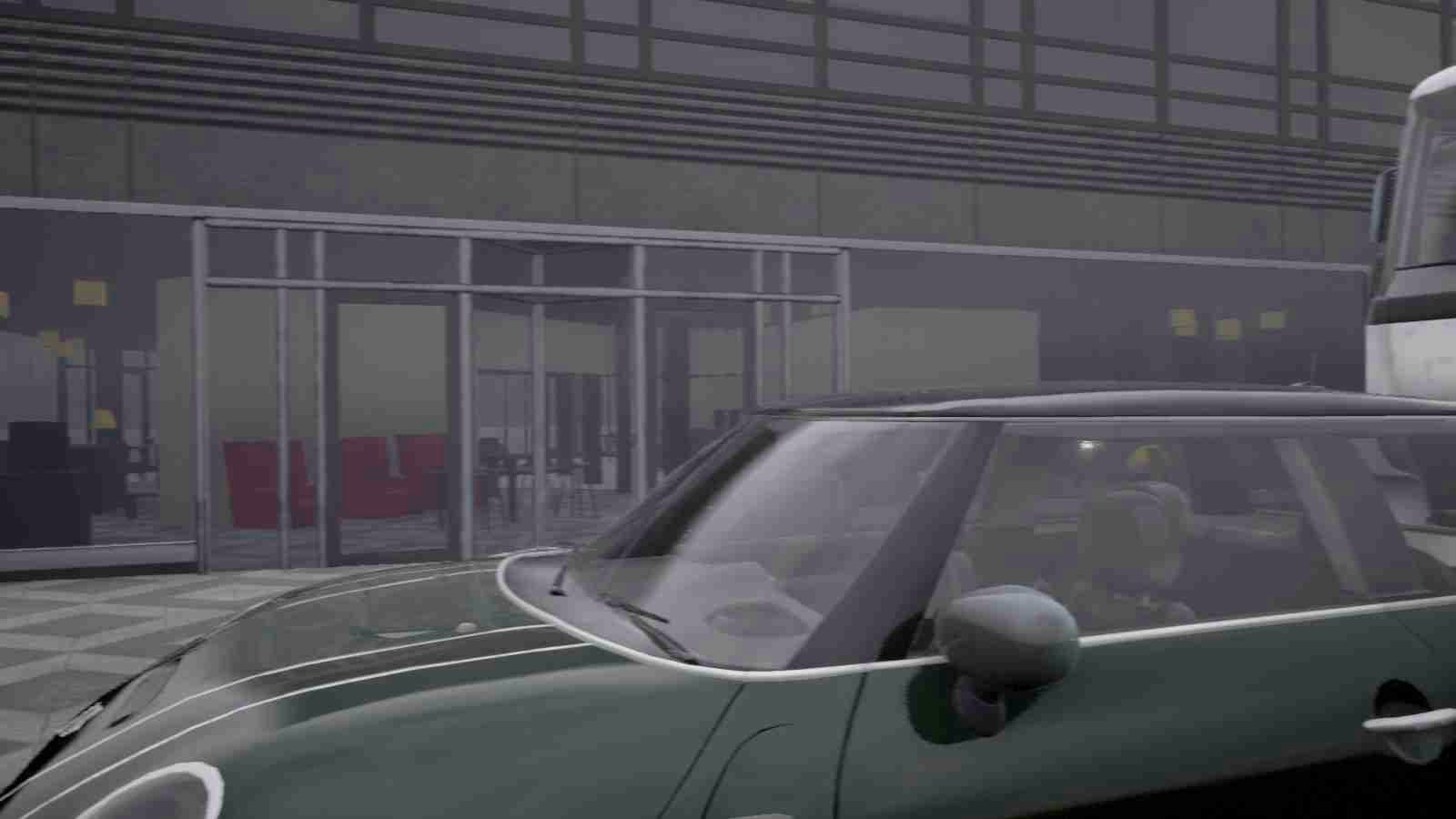}
        \caption{Back right}
    \end{subfigure}

    \vspace{-10pt}
    \caption{Six camera images of $CAV_{c}$ at frame $t$ that generates the BEV occupancy map $O_{c}$.}
    \label{fig:communicating_six_camera_images}
\end{figure*}

\begin{figure*}[t]
    \centering
    \begin{subfigure}[t]{0.34\textwidth}
        \centering
        \includegraphics[width=\linewidth]{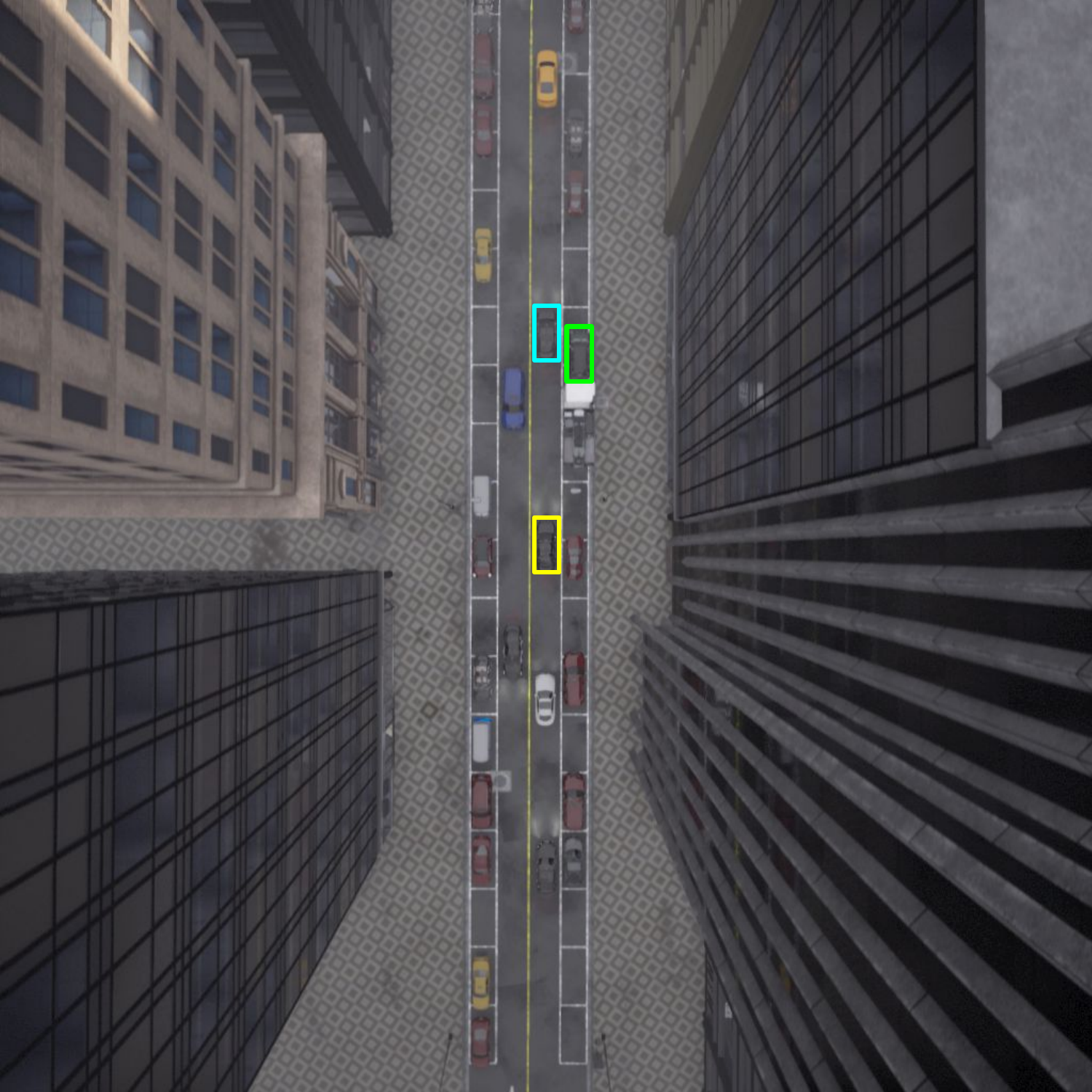}
        \caption{BEV camera view centered at $CAV_{ego}$. Only used for visualization in the paper, not used in model input or output.}
        \label{fig:bev_camera_cav_ego}
    \end{subfigure}%
    \hspace{0.02\textwidth}
    \begin{subfigure}[t]{0.34\textwidth}
        \centering
        \includegraphics[width=\linewidth]{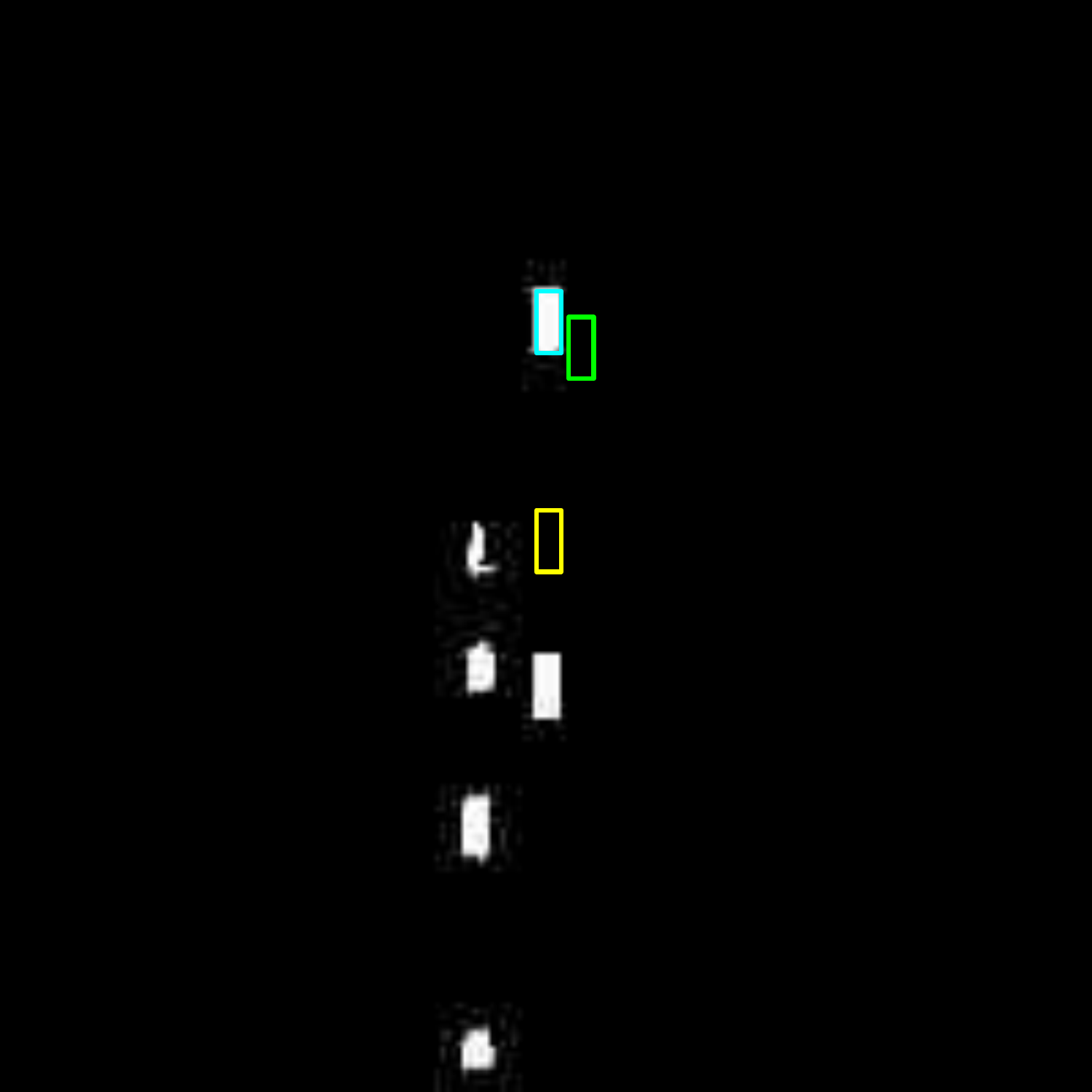}
        \caption{$O_{ego}$: BEV occupancy map from $CAV_{ego}$ in the coordinate system of $CAV_{ego}$.}
        \label{fig:bev_cav_0}
    \end{subfigure}


    \begin{subfigure}[t]{0.34\textwidth}
        \centering
        \includegraphics[width=\linewidth]{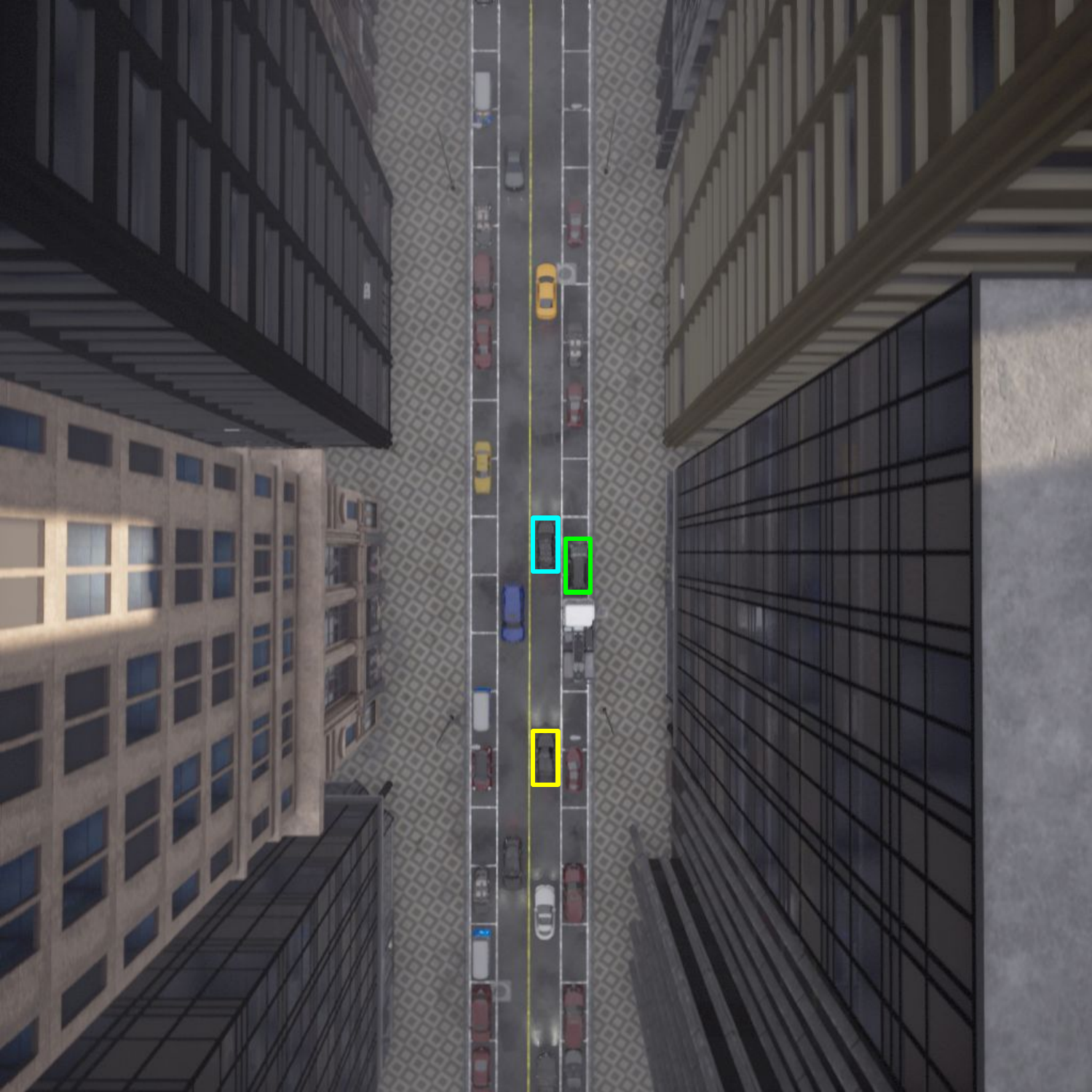}
        \caption{BEV camera view centered at $CAV_{c}$. Only used for visualization in the paper, not used in model input or output.}
        \label{fig:bev_camera_cav_communicating}
    \end{subfigure}%
    \hspace{0.02\textwidth}
    \begin{subfigure}[t]{0.34\textwidth}
        \centering
        \includegraphics[width=\linewidth]{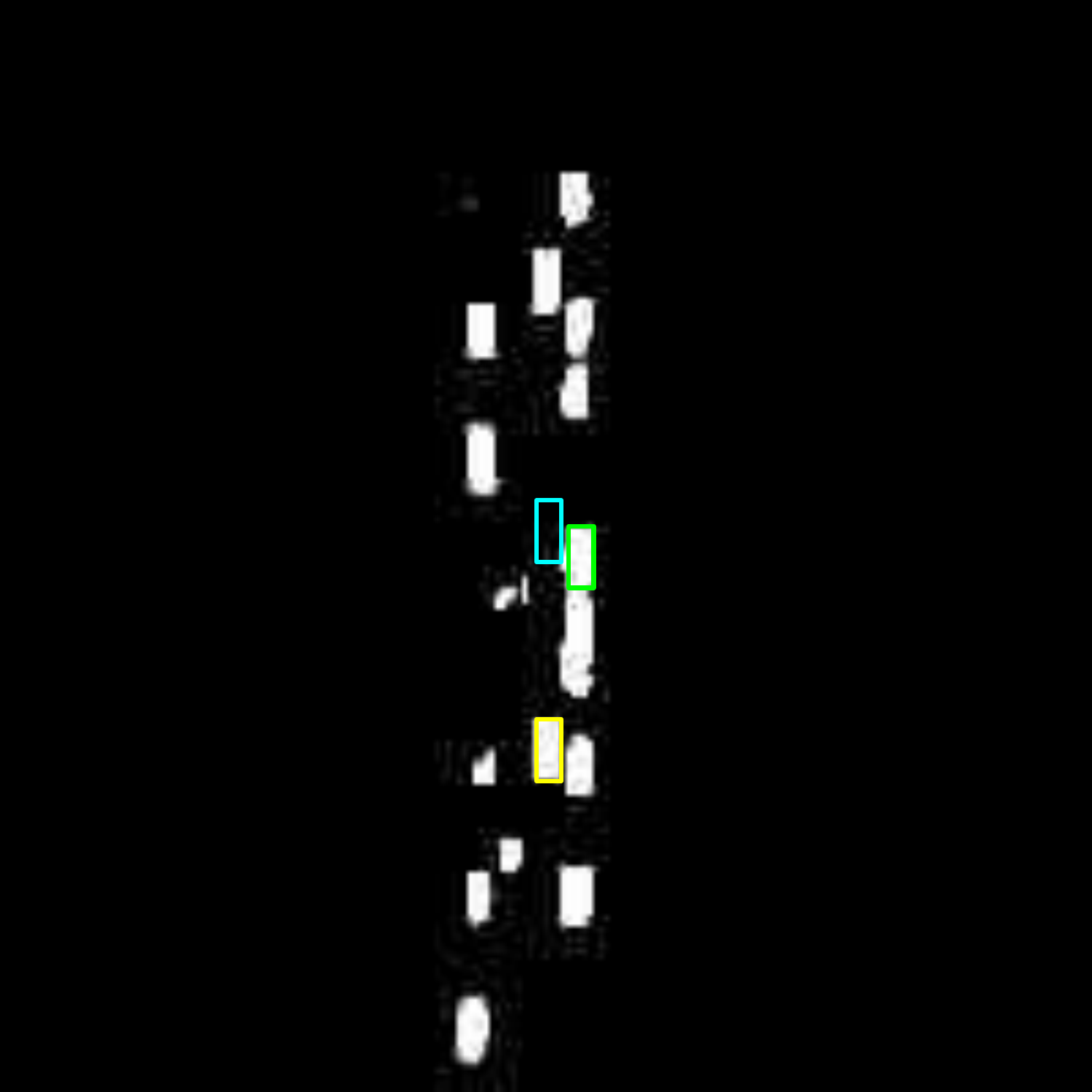}
        \caption{$O_{c}$: BEV occupancy map from $CAV_{c}$ in the coordinate system of $CAV_{c}$.}
        \label{fig:bev_cav_2_before_warp}
    \end{subfigure}


    \begin{subfigure}[t]{0.34\textwidth}
        \centering
        \includegraphics[width=\linewidth]{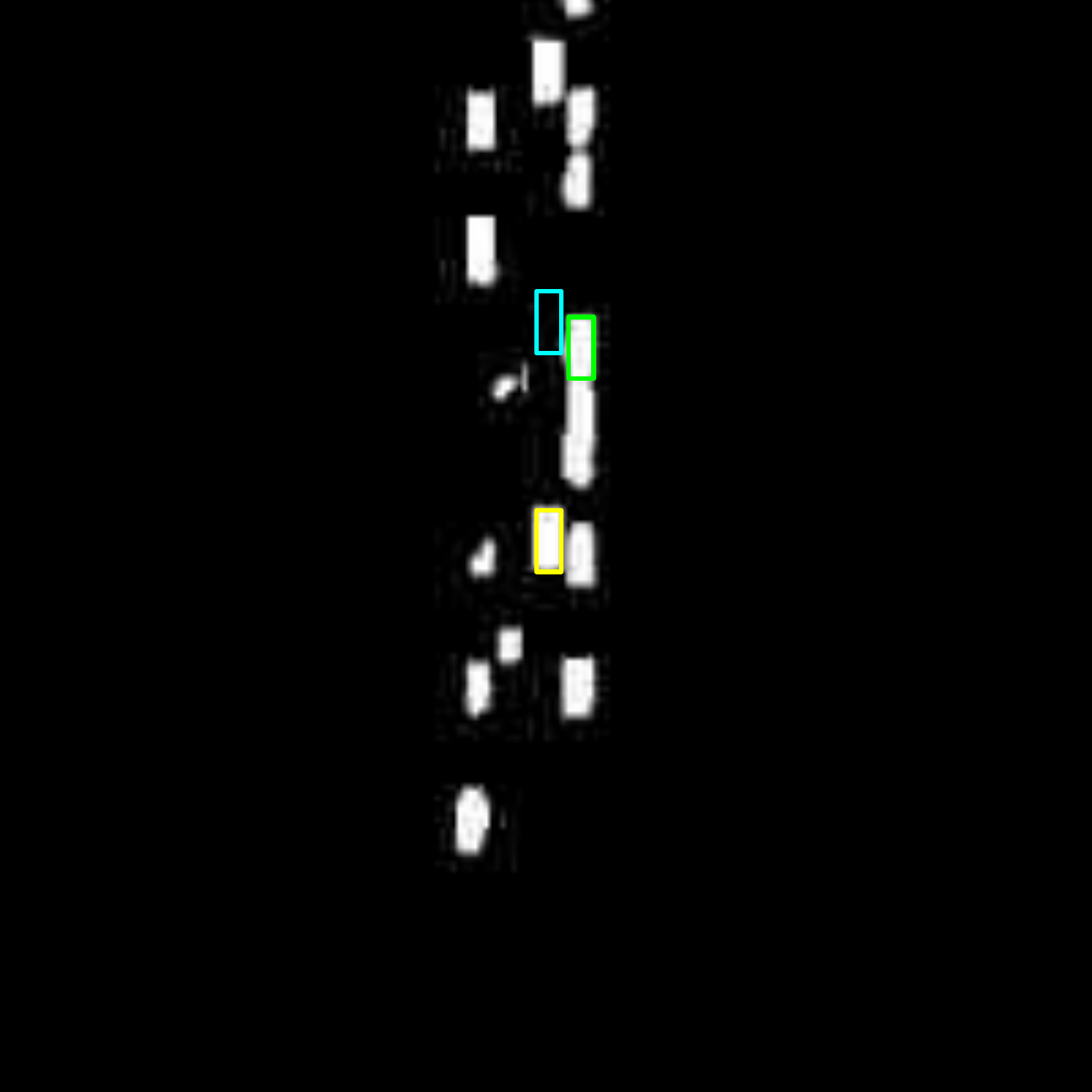}
        \caption{$O_{c -> ego}$: BEV occupancy map from $CAV_{c}$ in the coordinate system of $CAV_{ego}$.}
        \label{fig:bev_cav_2_after_warp}
    \end{subfigure}%
    \hspace{0.02\textwidth}
    \begin{subfigure}[t]{0.34\textwidth}
        \centering
        \includegraphics[width=\linewidth]{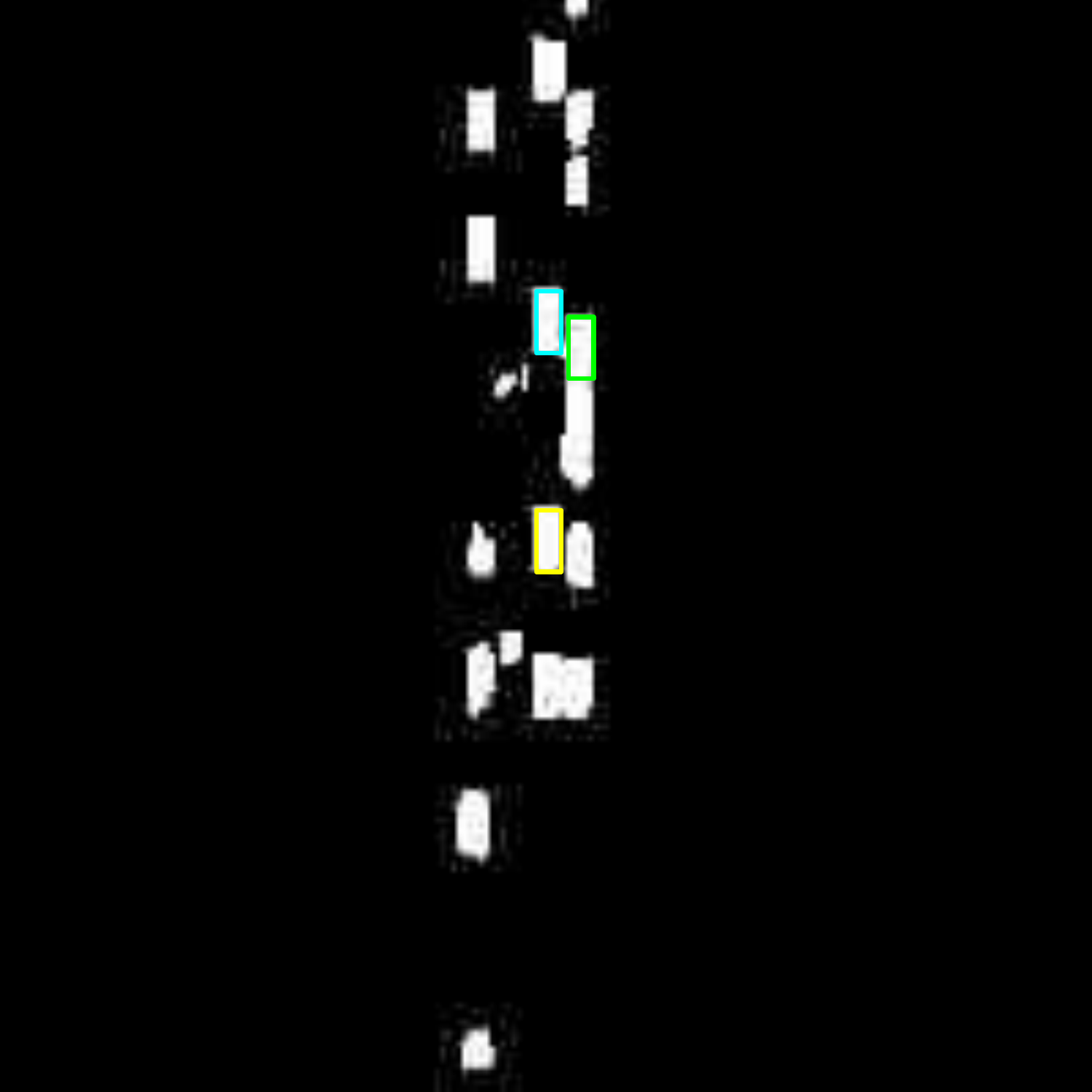}
        \caption{$O_{merged}$: Merged BEV occupancy map in the coordinate system of $CAV_{ego}$.}
        \label{fig:bev_merged}
    \end{subfigure}

    \vspace{-10pt} 
    \caption{BEV occupancy maps at frame $t$. \textcolor{darkyellow}{Yellow} box: $CAV_{ego}$. \textcolor{cyan}{Cyan} box: $CAV_{c}$. \textcolor{darkgreen}{Green} box: the critical conflicting green vehicle. The ego car $CAV_{ego}$ is unable to detect the critical green vehicle, while its selected communicative vehicle $CAV_c$ is able to do so. The occupancy map $O_c$ from $CAV_c$ is shared with $CAV_{ego}$ and helps $CAV_{ego}$ to decide to decelerate and avoid the potential collision.}
    \label{fig:bev_occupancy_maps}
\end{figure*}

In our proposed \namemethod, the ego vehicle $CAV_{ego}$ first uses the UniAD~\cite{hu2023uniad} model and $CAV_{ego}$'s individual 6 camera images, as shown in \cref{fig:ego_six_camera_images}, to generate its individual Bird-Eye-View (BEV) occupancy map $O_{ego}$, as shown in \cref{fig:bev_cav_0}. Similarly, the selected communicating vehicle $CAV_c$  uses $CAV_{c}$'s individual 6 camera images, as shown in \cref{fig:communicating_six_camera_images}, to generate its individual BEV occupancy map $O_{c}$, as shown in \cref{fig:bev_cav_2_before_warp}.
Subsequently, $O_c$ is warped to generate $O_{c -> ego}$ via coordinate transformation based on the ego vehicle pose $P_{ego}$ and the communicating vehicle pose $P_c$, as shown in \cref{fig:bev_cav_2_after_warp}.
Finally, $O_{merged}$ is generated by merging $O_{ego}$ and $O_{c -> ego}$ with the element-wise-or operator, as shown in \cref{fig:bev_merged}. $O_{merged}$ is then rendered as an RGB image, and is fed into the vision encoder of \namemethod, together with the $CAV_{ego}$'s front camera image $I_{ego}$ to form the vision input $V_{input}$. 

From the camera images and BEV occupancy maps shown in \cref{fig:ego_six_camera_images,fig:communicating_six_camera_images,fig:bev_occupancy_maps}, we can observe that the ego car $CAV_{ego}$ is unable to detect the green vehicle occluded by the white front right parked truck, while that green vehicle is cutting into the driving lane starting from the frame $t$. However, the selected communicative vehicle $CAV_{c}$ in front is able to detect the green vehicle and shares its BEV occupancy map $O_c$ with the ego car $CAV_{ego}$. This shared perception information helps the ego car $CAV_{ego}$ to decelerate even before the green vehicle is visible on the camera of $CAV_{ego}$ and avoid potential collisions.

\section{More Qualitative Evaluation Results}
More results can be seen in the supplementary video.

\clearpage

\begin{table*}[tb]
  \caption{Driving scenarios and skill categorization in \namedataset. \namedataset~has 50 types of safety-critical driving scenarios. Each scenario is instantiated under 5 different weather and lighting conditions, resulting in a total of 250 evaluation routes. Each scenario is designed to evaluate one or more different driving skills.
  }
  \vspace{-10pt}
  \label{tab:evaluation_route_skill}
  \footnotesize
  \centering
  \begin{tabular}{l | c | c | l }
    \toprule
    Skill & \# Scenarios & \# Routes & Scenarios \\
    \midrule
    Merging & 17 & 85 & CrossingBicycleFlow, \\
    & & & 
    EnterActorFlow, \\
    & & & 
    HighwayCutIn, HighwayExit,  \\
    & & &
    InterurbanActorFlow, InterurbanAdvancedActorFlow, \\
    & & & 
    LargeSignalizedJunctionLeftTurn,   \\
    & & & 
    MergerIntoSlowTraffic, MergerIntoSlowTrafficV2,  \\
    & & & 
    NonSignalizedJunctionLeftTurn, NonSignalizedJunctionLeftTurnEnterFlow, \\
    & & &
    NonSignalizedJunctionRightTurn, \\
    & & &
    ParkingExit,  \\
    & & & 
    SequentialLaneChange, SignalizedJunctionLeftTurn, SignalizedJunctionLeftTurnEnterFlow, \\
    & & &
    SignalizedJunctionRightTurn\\
    \midrule
    Overtaking & 9 & 45 & Accident, AccidentTwoWays, \\
    & & & 
    ConstructionObstacle, ConstructionObstacleTwoWays, \\
    & & & 
    HazardAtSideLane, HazardAtSideLaneTwoWays, \\
    & & & 
    ParkedObstacle, ParkedObstacleTwoWays, \\
    & & & 
    VehicleOpensDoorTwoWays \\
    \midrule
    EmergencyBrake & 17 & 85 & BlockedIntersection, \\
    & & & 
    ControlLoss, \\
    & & &
    DynamicObjectCrossing, \\
    & & & 
    HardBreakRoute,  \\
    & & &
    LargeParkingCrossingPedestrian, LargeParkingCutIn, LargeStaticCutIn, \\
    & & &
    LargeVehicleTurningRoute, LargeVehicleTurningRoutePedestrian, \\
    & & & 
    OppositeVehicleRunningRedLight, OppositeVehicleTakingPriority, \\
    & & & 
    ParkingCrossingPedestrian, ParkingCutIn, PedestrianCrossing, \\
    & & &
    StaticCutIn, \\ 
    & & &
    VehicleTurningRoute, VehicleTurningRoutePedestrian  \\
    \midrule
    GiveWay & 2 & 10 & InvadingTurn, \\
    & & & 
    YieldToEmergencyVehicle \\
    \midrule
    TrafficSign & 21 & 105 & CrossingBicycleFlow, \\
    & & & 
    EnterActorFlow, \\
    & & &
    LargeSignalizedJunctionLeftTurn, LargeVehicleTurningRoute, LargeVehicleTurningRoutePedestrian, \\
    & & &
    NonSignalizedJunctionLeftTurn, NonSignalizedJunctionLeftTurnEnterFlow, \\
    & & &
    NonSignalizedJunctionRightTurn, \\
    & & & 
    OppositeVehicleRunningRedLight, OppositeVehicleTakingPriority, \\
    & & & 
    PedestrianCrossing, \\
    & & & 
    SignalizedJunctionLeftTurn, SignalizedJunctionLeftTurnEnterFlow, \\
    & & &
    SignalizedJunctionRightTurn, \\
    & & &
    T\_Junction, \\
    & & & 
    VanillaNonSignalizedTurn, VanillaNonSignalizedTurnEncounterStopsign, \\
    & & &  
    VanillaSignalizedTurnEncounterGreenLight, VanillaSignalizedTurnEncounterRedLight, \\
    & & &
    VehicleTurningRoute, VehicleTurningRoutePedestrian \\
  \bottomrule
  \end{tabular}
\end{table*}

\clearpage
\begin{table*}[tb]
  \caption{Description of the first half $25$ driving scenarios out of the total $50$ scenarios in \namedataset. The descriptions include information directly taken or updated from Bench2Drive~\cite{jia2024bench} and new information in \namedataset.
  }
  \vspace{-10pt}
  \label{tab:scenraio_description_1}
  \footnotesize
  \centering
  \begin{tabular}{r | l | l}
    \toprule
    ID & Scenarios & Description \\
    \midrule
    1 & Accident & The ego vehicle encounters multiple accident cars blocking part of the lane and must perform \\
    & & 
    a lane change into traffic in the same direction. \\
    \midrule
    2 & AccidentTwoWays & The ego vehicle encounters multiple accident cars blocking part of the lane and must perform \\
    & & 
    a lane change into traffic in the opposite direction. \\
    \midrule
    3 & BlockedIntersection & While performing a maneuver, the ego vehicle encounters a stopped vehicle on the road and \\
    & & 
    must perform an emergency brake or an avoidance maneuver. \\
    \midrule
    4 & ConstructionObstacle & The ego vehicle encounters a construction site blocking and must perform a lane change into \\
    & & 
    traffic in the same direction. \\
    \midrule
    5 & ConstructionObstacleTwoWays & The ego vehicle encounters a construction site blocking and must perform a lane change \\
    & & 
    into traffic in the opposite direction. \\
    \midrule
    6 & ControlLoss & The ego vehicle loses control due to bad conditions on the road and it must recover, \\
    & & 
    coming back to its original lane. \\
    \midrule
    7 & CrossingBicycleFlow & The ego vehicle needs to perform a turn at an intersection yielding to bicycles crossing from \\
    & & 
    either the left or right. \\
    \midrule
    8 & DynamicObjectCrossing & A walker or bicycle behind a static prop crosses the road suddenly when the ego vehicle is close \\
    & & 
    to the prop. The ego vehicle must make a hard brake promptly. \\
    \midrule
    9 & EnterActorFlow & A flow of cars runs a red light in front of the ego when it enters the junction, forcing it to react. \\
    \midrule
    10 & HardBreakRoute & The leading vehicle decelerates suddenly and the ego vehicle must perform an emergency brake \\
    & & 
    or an avoidance maneuver. \\
    \midrule
    11 & HazardAtSideLane & The ego vehicle encounters a slow-moving hazard blocking part of the lane. The ego vehicle \\
    & & 
    must brake or maneuver next to a lane of traffic in the same direction. \\
    \midrule
    12 & HazardAtSideLaneTwoWays & The ego vehicle encounters a slow-moving hazard blocking part of the lane. The ego vehicle \\
    & & 
    must brake or maneuver next to a lane of traffic in the opposite direction. \\
    \midrule
    13 & HighwayCutIn & The ego vehicle encounters a vehicle merging into its lane from a highway on-ramp. The ego \\
    & & 
    vehicle must decelerate, brake, or change lanes to avoid a collision. \\
    \midrule
    14 & HighwayExit & The ego vehicle must cross a lane of moving traffic to exit the highway at an off-ramp. \\
    \midrule
    15 & InterurbanActorFlow & The ego vehicle leaves the interurban road by turning left, crossing a fast traffic flow. \\
    \midrule
    16 & InterurbanAdvancedActorFlow & The ego vehicle incorporates into the interurban road by turning left, first crossing a fast traffic \\
    & & 
    flow, and then merging into another one. \\
    \midrule
    17 & InvadingTurn & When the ego vehicle is about to turn right, a vehicle coming from the opposite lane invades the \\
    & & 
    ego’s lane, forcing the ego to move right to avoid a possible collision. \\
    \midrule
    18 & LargeParkingCrossingPedestrian & The ego vehicle encounters a pedestrian emerging from behind a parked bus or truck and \\
    & & 
    advancing into the lane. The ego vehicle must brake or maneuver to avoid it. \\
    \midrule
    19 & LargeParkingCutIn & The ego vehicle must slow down or brake to allow a parked vehicle behind a parked bus or truck \\
    & & 
    exiting a parallel parking bay to cut in front. \\
    \midrule
    20 & LargeSignalizedJunctionLeftTurn & The ego vehicle is performing an unprotected left turn at an signaled intersection, yielding to \\
    & & 
    oncoming traffic of buses or trucks. \\
    \midrule
    21 & LargeStaticCutIn & The ego vehicle must slow down or brake to allow a vehicle in the slow traffic flow of buses or \\
    & & 
    trucks in the adjacent lane to cut in front. \\
    \midrule
    22 & LargeVehicleTurningRoute & While performing a maneuver, the ego vehicle encounters a bicycle crossing the road with \\
    & & 
    background traffic of buses or trucks, and must perform an emergency brake or an avoidance \\
    & & maneuver. \\
    \midrule
    23 & LargeVehicleTurningRoutePedestrian & While performing a maneuver, the ego vehicle encounters a pedestrian crossing the road with \\
    & & 
    background traffic of buses or trucks, and must perform an emergency brake or an avoidance \\
    & & 
    maneuver. \\
    \midrule
    24 & MergerIntoSlowTraffic & The ego vehicle must merge into a slow traffic flow on the off-ramp when exiting the highway. \\
    \midrule
    25 & MergerIntoSlowTrafficV2 & The ego vehicle must merge into a slow traffic flow coming from the on-ramp when driving on \\
    & & highway roads. \\
  \bottomrule
  \end{tabular}
\end{table*}

\clearpage
\begin{table*}[tb]
  \caption{Description of the second half $25$ driving scenarios out of the total $50$ scenarios in \namedataset. The descriptions include information   directly taken or updated from Bench2Drive~\cite{jia2024bench} and new information in \namedataset.
  }
  \vspace{-10pt}
  \label{tab:scenraio_description_2}
  \footnotesize
  \centering
  \begin{tabular}{r | l | l}
    \toprule
    ID & Scenarios & Description \\
    \midrule
    26 & NonSignalizedJunctionLeftTurn & The ego vehicle is performing an unprotected left turn at a non-signaled intersection, yielding  \\
    & & 
    to oncoming traffic. \\
    \midrule
    27 & NonSignalizedJunctionLeftTurnEnterFlow & The ego vehicle is performing an unprotected left turn at a non-signaled intersection, merging \\
    & & 
    into opposite traffic. \\
    \midrule
    28 & NonSignalizedJunctionRightTurn & The ego vehicle is turning right at a non-signaled intersection and has to safely merge into the \\
    & & 
    traffic flow coming from its left. \\
    \midrule
    29 & OppositeVehicleRunningRedLight & The ego vehicle is going straight at an intersection but a crossing vehicle runs a red light,  \\
    & & 
    forcing the ego vehicle to avoid the collision. \\
    \midrule
    30 & OppositeVehicleTakingPriority & The ego vehicle is going straight at a non-signaled intersection but a crossing vehicle enters the \\
    & &
    intersection at the same time without slowing down, forcing the ego vehicle to avoid the \\
    & & 
    collision. \\
    \midrule
    31 & ParkedObstacle & The ego vehicle encounters a parked vehicle blocking part of the lane and must perform a lane \\
    & & 
    change into traffic in the same direction. \\
    \midrule
    32 & ParkedObstacleTwoWays & The ego vehicle encounters a parked vehicle blocking part of the lane and must perform a lane \\
    & & 
    change into traffic in the opposite direction. \\
    \midrule
    33 & ParkingCrossingPedestrian & The ego vehicle encounters a pedestrian emerging from behind a parked vehicle and advancing \\
    & & 
    into the lane. The ego vehicle must brake or maneuver to avoid it. \\
    \midrule
    34 & ParkingCutIn & The ego vehicle must slow down or brake to allow a parked vehicle exiting a parallel parking  \\
    & & 
    bay to cut in front. \\
    \midrule
    35 & ParkingExit & The ego vehicle must exit a parallel parking bay into a flow of traffic. \\
    \midrule
    36 & PedestrianCrossing & While the ego vehicle is entering a junction, a group of natural pedestrians suddenly cross the  \\
    & & 
    road and ignore the traffic light. The ego vehicle must stop and wait for all pedestrians to pass \\
    & & 
    even though there is a green traffic light or a clear junction. \\
    \midrule
    37 & SequentialLaneChange & A basic scenario for the ego vehicle to learn to change multiple lanes and avoid collision. \\
    \midrule
    38 & SignalizedJunctionLeftTurn & The ego vehicle is performing an unprotected left turn at an intersection, yielding to \\
    & & 
    oncoming traffic. \\
    \midrule
    39 & SignalizedJunctionLeftTurnEnterFlow & The ego vehicle is performing an unprotected left turn at an intersection, merging into \\
    & & 
    opposite traffic. \\
    \midrule
    40 & SignalizedJunctionRightTurn & The ego vehicle is turning right at an intersection and has to safely merge into the traffic flow \\
    & & 
    coming from its left. \\
    \midrule
    41 & StaticCutIn & The ego vehicle must slow down or brake to allow a vehicle of the slow traffic flow in the  \\
    & & 
    adjacent lane to cut in front. \\
    \midrule
    42 & T\_Junction & A basic scenario for the ego vehicle to learn to pass through a T-junction. \\
    \midrule
    43 & VanillaNonSignalizedTurn & A basic scenario for the ego vehicle to learn to pass through a non-signalized junction. \\
    \midrule
    44 & VanillaNonSignalizedTurnEncounterStopsign & A basic scenario for the ego vehicle to learn to stop and start at stop signs. \\
    \midrule
    45 & VanillaSignalizedTurnEncounterGreenLight & A basic scenario for the ego vehicle to learn to pass through a signalized junction. \\
    \midrule
    46 & VanillaSignalizedTurnEncounterRedLight & A basic scenario for the ego vehicle to learn to pass through a signalized junction when the  \\
    & & 
    traffic light changes from red to green. \\
    \midrule
    47 & VehicleOpensDoorTwoWays & The ego vehicle encounters a parked vehicle opening a door into its lane and must maneuver to \\
    & & 
    avoid it. \\
    \midrule
    48 & VehicleTurningRoute & While performing a maneuver, the ego vehicle encounters a bicycle crossing the road and must \\
    & & 
    perform an emergency brake or an avoidance maneuver. \\
    \midrule
    49 & VehicleTurningRoutePedestrian & While performing a maneuver, the ego vehicle encounters a pedestrian crossing the road and  \\
    & & 
    must perform an emergency brake or an avoidance maneuver. \\
    \midrule
    50 & YieldToEmergencyVehicle & The ego vehicle is approached by an emergency vehicle coming from behind. The ego vehicle  \\
    & & 
    must maneuver to allow the emergency vehicle to pass. \\
  \bottomrule
  \end{tabular}
\end{table*}

\end{document}